%% file: iclr2026_conference.tex
\documentclass{article} 
\usepackage{iclr2026_conference,times}

\input{math_commands.tex}

\usepackage{url}
\usepackage{booktabs}       
\usepackage{xcolor}         
\usepackage{graphicx}
\usepackage{adjustbox}
\usepackage{wrapfig}
\usepackage{caption}
\usepackage{makecell}
\usepackage{multirow}
\usepackage{amsmath}
\usepackage{hyperref}
\makeatletter
  \newcommand\figcaption{\def\@captype{figure}\caption}
  \newcommand\tabcaption{\def\@captype{table}\caption}
\makeatother

\title{OCR-Reasoning Benchmark: Unveiling the True Capabilities of MLLMs in Complex Text-Rich Image Reasoning}

\author{Mingxin Huang$^{\dag1}$, Yongxin Shi$^{\dag1}$, Dezhi Peng$^{*2}$, Songxuan Lai$^{2}$, Zecheng Xie$^{2}$, Lianwen Jin$^{*1}$ \\
$^1$South China University of Technology  $^2$Huawei Technologies Co., Ltd.
}

\iclrfinalcopy 
\begin{document}

\renewcommand{\thefootnote}{\fnsymbol{footnote}}
{\let\thefootnote\relax\footnotetext{
\noindent \hspace{-5mm}$^\dag$Equal contribution. \\
$^*$Corresponding authors.}}

\maketitle

\begin{abstract}
Recent advancements in multimodal slow-thinking systems have demonstrated remarkable performance across various visual reasoning tasks. However, their capabilities in text-rich image reasoning tasks remain understudied due to the absence of a dedicated and systematic benchmark. To address this gap, we propose \textbf{OCR-Reasoning}, a novel benchmark designed to systematically assess Multimodal Large Language Models on text-rich image reasoning tasks. Specifically, OCR-Reasoning comprises 1,069 human-annotated examples spanning 6 core reasoning abilities and 18 practical reasoning tasks in text-rich visual scenarios. Unlike existing text-rich image understanding benchmarks that only provide a final answer, this benchmark additionally provides a detailed step-by-step reasoning process. This dual annotation enables the evaluation of both the models' final answers and their reasoning processes, thereby offering a holistic assessment of text-rich reasoning capabilities. By leveraging this benchmark, we conducted a comprehensive evaluation of the latest MLLMs. Our results demonstrate that even the most advanced MLLMs exhibit substantial difficulties in text-rich image reasoning tasks, with none achieving an accuracy above 50\% on our benchmark, indicating that the challenges of text-rich image reasoning are an urgent issue to be addressed. The benchmark and evaluation scripts are available at https://github.com/SCUT-DLVCLab/OCR-Reasoning.
\end{abstract}

\section{Introduction}

Recently, slow-thinking systems in Large Language Models (LLMs), such as OpenAI-o1~\citep{jaech2024openai}, DeepSeek-R1~\citep{guo2025deepseek}, Gemini-Thinking~\citep{team2023gemini}, and QwQ~\citep{qwq32b} have demonstrated significant progress in addressing complex math, coding, logical, and scientific problems. Building upon techniques like Chain-of-Thought (CoT) prompting~\citep{wei2022chain} and test-time compute scaling~\citep{jaech2024openai,guo2025deepseek}, slow-thinking systems typically engage in critical thinking and reflection before providing the final answer. Moreover, emerging evidence suggests these systems may even experience `Aha moments' when solving complex problems~\citep{guo2025deepseek}. In order to broaden their ability across diverse contexts, multimodal slow-thinking systems have emerged as a rapidly evolving research direction, driven by the need for more versatile AI applications~\citep{yang2025r1onevision, peng2025lmmr1, meng2025mm, chen2025r1v, liu2025visual, wang2025visualprm, liu2025othink, shen2025vlmr1,vl-rethinker,liu2025noisyrollout}. 

To assess the reasoning capabilities of multimodal slow-thinking systems, researchers have developed specialized reasoning benchmarks targeting distinct scenarios. For instance, MathVista~\citep{lu2023mathvista}, MathVerse~\citep{zhang2024mathverse}, Olympiadbench~\citep{he2024olympiadbench}, and MathVision~\citep{wang2024measuring} are designed to evaluate the math-related reasoning ability of the model. In college-level subject knowledge domains, MMMU~\citep{yue2023mmmu} focuses on advanced reasoning in domains such as chemistry, physics, and scientific problem-solving. While these domains are thriving with the corresponding benchmarks, a critical gap persists in text-rich image scenarios. Current benchmarks for text-rich images, such as DocVQA~\citep{mathew2021docvqa}, ChartQA~\citep{masry2022chartqa}, and OCRBench~\citep{liu2024ocrbench}, are designed primarily to assess the ability to merely extract textual content without requiring in-depth analysis~\citep{mathew2021docvqa,xu2025visulogic}. However, text-rich images involve many reasoning-intensive tasks such as financial report analysis, invoice analysis, and cost-effective purchase decisions~\citep{gan2024mme,sun2021spatial}. There is still a lack of benchmarks for systematically evaluating the reasoning ability within text-rich visual scenarios.

\begin{figure}[t!]
\centering
    \includegraphics[width=0.9\linewidth]{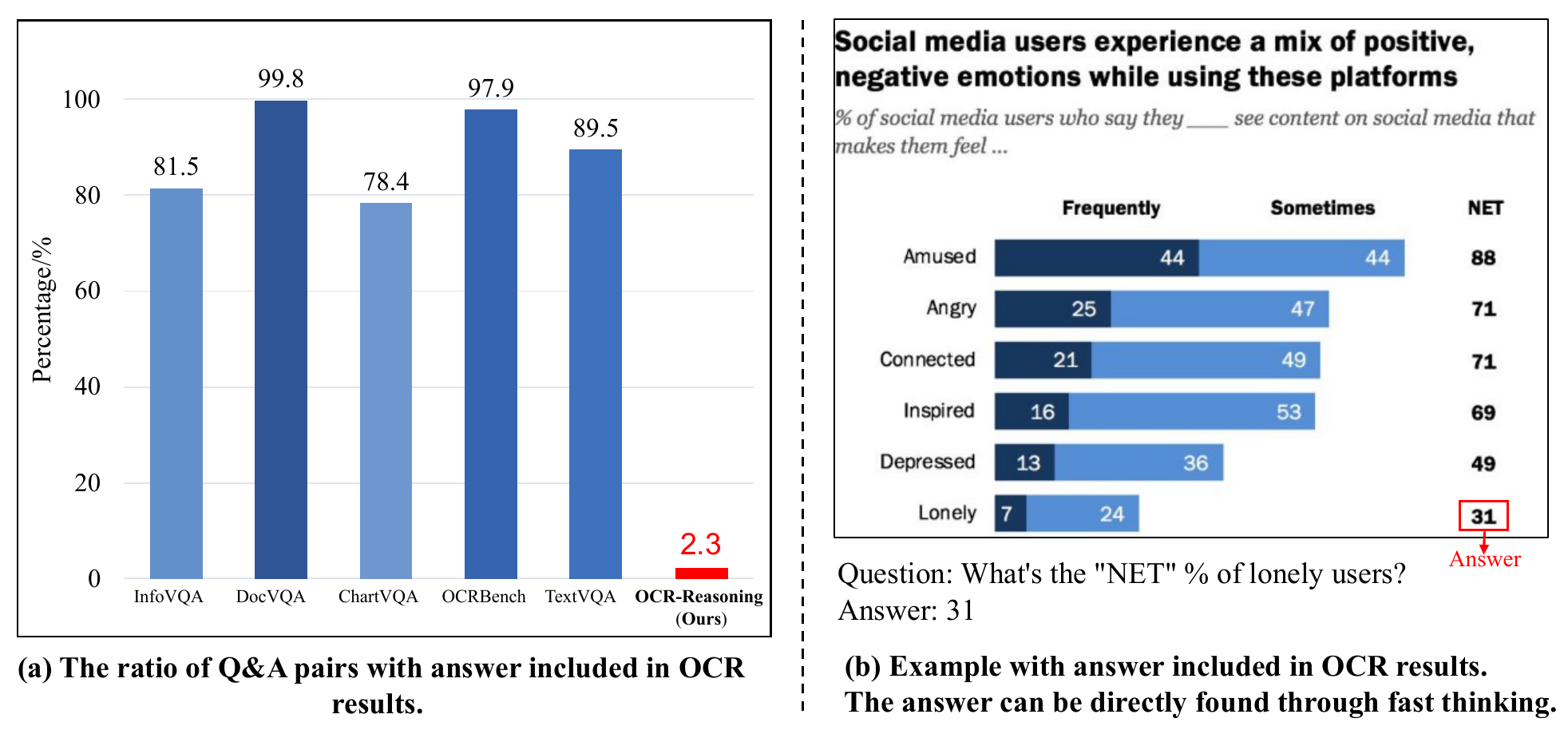}
    \caption{
        (a) The percentage of answers in the benchmark's $Q\&A$ pairs that can be retrieved from the OCR results. (b) An example where the answers can be retrieved from the OCR results.
    }
    \label{fig:VIE_percent}
\end{figure}

To bridge this critical gap in multimodal evaluation, we introduce OCR-Reasoning, a novel benchmark designed to evaluate the text-rich image reasoning skills of Multimodal Large Language Models (MLLMs). Specifically, our benchmark contains 1,069 meticulously collected and human-annotated examples, which span 6 core reasoning abilities and 18 practical reasoning tasks commonly found in text-rich visual contexts. Furthermore, unlike other text-rich image understanding benchmarks that only annotate the final answers, OCR-Reasoning provides annotations for both the final answers and the step-by-step reasoning process. This comprehensive annotation scheme facilitates a more in-depth evaluation of MLLMs' reasoning capabilities. Additionally, through a simple comparison with existing benchmarks, as shown in Fig.~\ref{fig:VIE_percent}, we observe that in most cases the answers in existing datasets are directly present in the images, whereas our benchmark contains very few samples of this type. This implies that in our benchmark, to obtain the answer, the model needs to engage in reasoning rather than extracting it from the OCR results of the image.

Using the OCR-Reasoning benchmark, we conduct extensive experiments to assess the text-rich image reasoning capabilities of popular LLMs and MLLMs. For pure LLMs, we replaced images with their OCR results and used these as input. The results show relatively low accuracy, which indicates that text alone is insufficient for solving text-rich image reasoning tasks. For MLLMs, the strongest performer achieves only 46.8\% accuracy, with none surpassing 50\% on our benchmark. As for document-oriented MLLMs, their highest accuracy does not exceed 15\%. These findings demonstrate that existing models still have significant room for improvement in handling text-rich image reasoning tasks. Additionally, we find that most of the existing reinforcement learning methods perform poorly on text-rich image reasoning tasks. Designing reinforcement learning for text-rich image reasoning is a potential direction for enhancing text-rich image reasoning capabilities.

The main contributions of this work are summarized as follows.

\begin{itemize}
    \item We introduce OCR-Reasoning, a challenging rich-text image reasoning benchmark that provides a systematic evaluation framework for assessing the reasoning capabilities of MLLMs in text-rich scenarios. To the best of our knowledge, we are the first to concretely define various core sub-abilities for text-rich image reasoning and conduct systematic evaluations.


    \item We conduct a systematic evaluation of leading MLLMs. Our results indicate that: 1) For text-rich image reasoning tasks, pure OCR input cannot effectively replace image input; 2) Even the most leading MLLMs struggle with our proposed benchmark. Based on the experiment results, we find several potential directions for future improvements.
    
\end{itemize}

\section{Related Work}

\subsection{Multi-modal Benchmark}
Driven by innovations in slow-thinking systems in LLMs, the evaluation of reasoning capabilities in Multimodal Large Language Models (MLLMs) has become a highly focused and widely discussed topic~\citep{lu2023mathvista,zhang2024mathverse,wang2024measuring,yue2023mmmu}. Early benchmarks such as CLEVR~\citep{johnson2017clevr} and GQA~\citep{hudson2019gqa} pioneered the integration of compositional language-vision abstraction to assess visual reasoning in structured environments. Subsequent works expanded the evaluation of reasoning into diverse domains. For instance, ScienceQA~\citep{lu2022learn} introduces scientific multimodal reasoning requiring domain knowledge. Meanwhile, the emergence of benchmarks like MMMU~\citep{yue2023mmmu} further pushes the boundaries by requiring a college-level reasoning across disciplines like physics and art. With the development of test-time compute scaling~\citep{jaech2024openai,cui2025process}, mathematical benchmarks requiring complex reasoning processes to obtain the answer are emerging as critical benchmarks for evaluating the reasoning capabilities of MLLMs. For instance, MathVista~\citep{lu2023mathvista} systematically categorizes seven mathematical reasoning types through multimodal problem decomposition. MathVision~\citep{wang2024measuring} curates competition-level mathematical problems with authentic visual contexts. Mathverse~\citep{zhang2024mathverse} introduces a comprehensive multimodal benchmark specifically designed to assess the visual mathematical reasoning capabilities of MLLMs. Although these benchmarks have expanded the scope of evaluation to various domains, there is still a lack of systematic evaluation in the widely applied field of text-rich image understanding. The text-rich image encompasses numerous scenarios requiring reasoning, such as financial report analysis, invoice analysis, cost-effective purchase decisions, and more.

\subsection{Text-rich Image Understanding Benchmark}
The evolution of Multimodal Large Language Models (MLLMs) has driven corresponding advancements in text-rich image understanding benchmarks. Early benchmarks for text-rich image understanding predominantly focused on assessing the perception capabilities of MLLMs within individual scenarios, such as documents~\citep{mathew2021docvqa}, charts~\citep{masry2022chartqa}, infographic images~\citep{mathew2022infographicvqa}, and scene text~\citep{singh2019towards,biten2019scene}. In parallel, recent advancements in high-resolution image processing~\citep{ye2023ureader,li2024monkey,huang2024mini,hu2024mplug15,guan2025token,liu2024hrvda} and optimized computational efficiency~\citep{liu2024textmonkey,hu2024mplug25,zhang2025dockylin,yu2024texthawk2} have significantly improved the performance of these benchmarks. To address the growing need for holistic evaluation of MLLMs, a series of benchmarks with broader, more diverse, and complex scenarios have emerged~\citep{wadhawan2024contextual,li2024seed,liu2024mmc,liu2024ocrbench,liu2024focus,ouyang2024omnidocbench}. For instance, OCRBench~\citep{liu2024ocrbench}, CC-OCR~\citep{yang2024cc}, and OCRBenchv2~\citep{fu2024ocrbench} concentrate on assessing the perceptual capabilities of MLLMs across multiple domains, while OmniDocBench~\citep{ouyang2024omnidocbench} provides a comprehensive evaluation of PDF document parsing. However, despite these advancements, with the emergence of slow-thinking systems requiring deliberate reasoning, current benchmarks reveal two critical limitations: 1. Overemphasis on textual extraction tasks~\citep{mathew2021docvqa,xu2025visulogic}, which can be solved through fast-thinking processes; 2. Lack of systematic assessment of reasoning capabilities in text-rich image understanding. This progression highlights the pressing need for next-generation benchmarks to evaluate MLLMs' complex reasoning capacities in text-rich visual understanding. To address this limitation, we propose a comprehensive benchmark specifically designed to assess multimodal slow-thinking systems in complex text-rich image reasoning tasks.

\begin{figure}[t!]
\centering
    \includegraphics[width=0.8\linewidth]{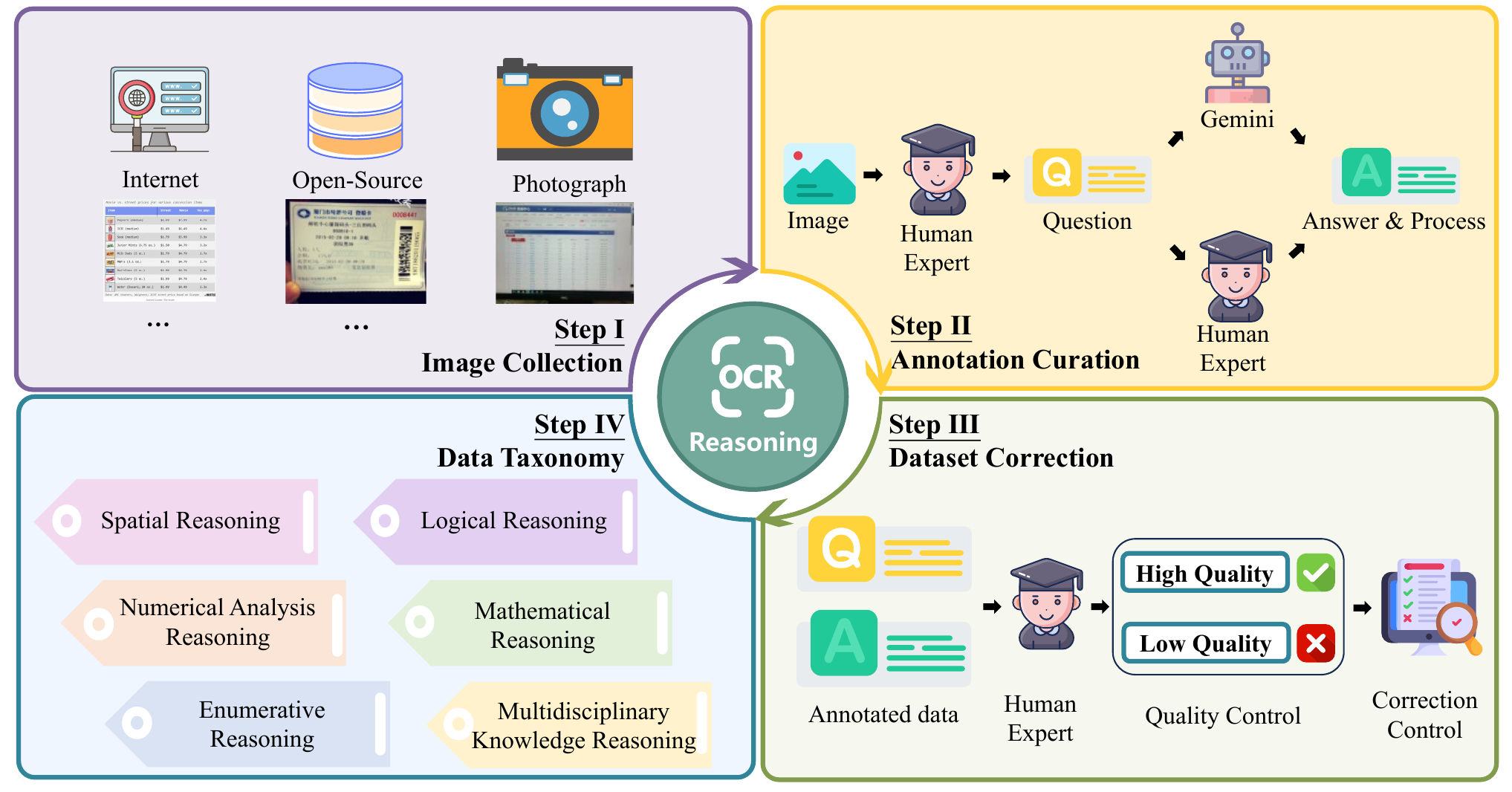}
    \caption{
        Data curation framework of OCR-Reasoning. The framework includes: (1) dataset collection, (2) annotation curation, (3) data correction, and (4) detailed taxonomy.
    }
    \label{fig:curation}
\end{figure}

\section{OCR-Reasoning}
In Sec.~\ref{sec:curation}, we first present the data curation framework of OCR-Reasoning, comprising: (1) dataset collection, (2) annotation curation, (3) data correction, and (4) detailed taxonomy. The data curation framework is shown in Fig.~\ref{fig:curation}. Then, in Sec.~\ref{sec:statistics}, we describe the statistics of OCR-Reasoning, including its total scale, categorical distribution, and detailed question-answer characteristics. Notably, while existing benchmarks~\citep{mathew2022infographicvqa,masry2022chartqa,liu2024ocrbench} focus solely on final answers, OCR-Reasoning provides annotations for both the final answers and the step-by-step reasoning process, facilitating a more in-depth evaluation of MLLMs' reasoning capabilities. The statistics of the annotations are presented in Sec.~\ref{sec:statistics}.

Additionally, OCR-Reasoning focuses on challenges in single-image. This design choice is based on two well-founded design principles: 1. Capability isolation and focused evaluation: Multi-image or multi-document tasks primarily assess long-context processing capabilities, which require specialized benchmarks for proper evaluation. Mixing single-image reasoning with multi-document challenges would confound the evaluation and make it difficult to isolate specific reasoning deficiencies.
2. Model compatibility and fair evaluation: Several document-oriented MLLMs~\citep{wang2025marten,xiao2025adaptive,guan2025token} only focus on single images and have not been trained on multiple images. Including multiple images in the benchmark would exclude these important models from evaluation, potentially reducing the focus on the reasoning problem itself.

\begin{figure}[t!]
\centering
    \includegraphics[width=\linewidth]{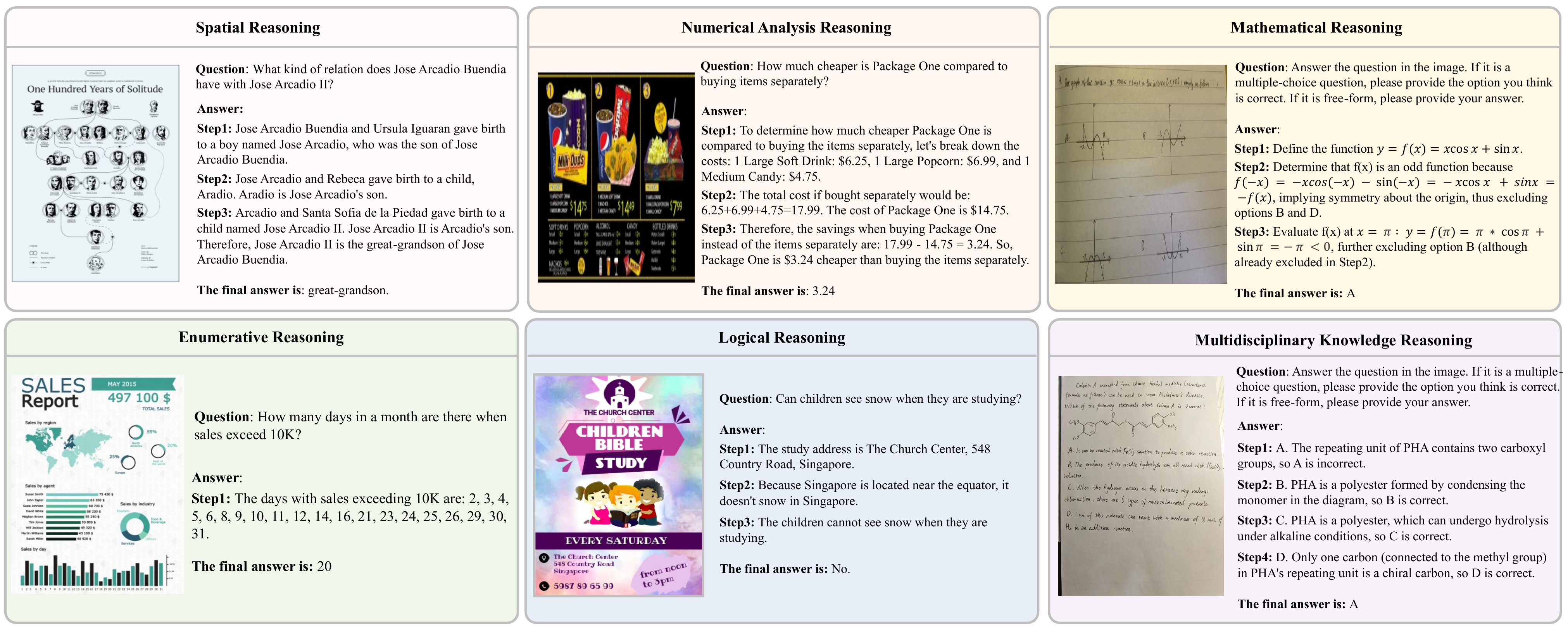}
    \caption{
        Examples of different categories in OCR-Reasoning. OCR-Reasoning includes six categories: spatial Reasoning, numerical analysis reasoning, mathematical reasoning, enumerative reasoning, logical reasoning, and multidisciplinary knowledge reasoning.
    }
    \label{fig:example}
\end{figure}

\begin{figure*}[t]
\centering
\begin{minipage}[c]{0.35\textwidth}
\small
\centering

  \label{t1}
  \centering
  \tabcaption{Key Statistics of OCR-Reasoning.}
  \begin{adjustbox}{width=\linewidth}
   \begin{tabular}{lr}
 \toprule
 \textbf{Statistic} & \textbf{Number} \\
 \midrule
  Total questions & 1069 \\
  ~- Multiple-choice questions & 250 (23.4\%) \\
  ~- Free-form questions & 819 (76.6\%) \\
  ~- Newly collected question & 987 (92.3\%) \\
  ~- Newly collected reasoning path & 1069 (100.0\%) \\
  \midrule
 Number of unique images & 1022 \\
 Number of unique questions & 1069 \\
 Number of unique answers & 1069 \\
 \midrule
 Maximum question length & 393 \\
 Maximum answer length & 3106 \\
 Average question length & 76\\
 Average answer length & 421\\
 \bottomrule
 \label{tab:statistics}
 \end{tabular}
 \end{adjustbox}
\end{minipage}
\qquad
\begin{minipage}[c]{0.52\textwidth}
\centering
\vspace{-0.2cm}
\caption{Subject Distribution of OCR-Reasoning.}
\label{fig3.5}
\vspace{0.15cm}
\includegraphics[width=0.88\linewidth]{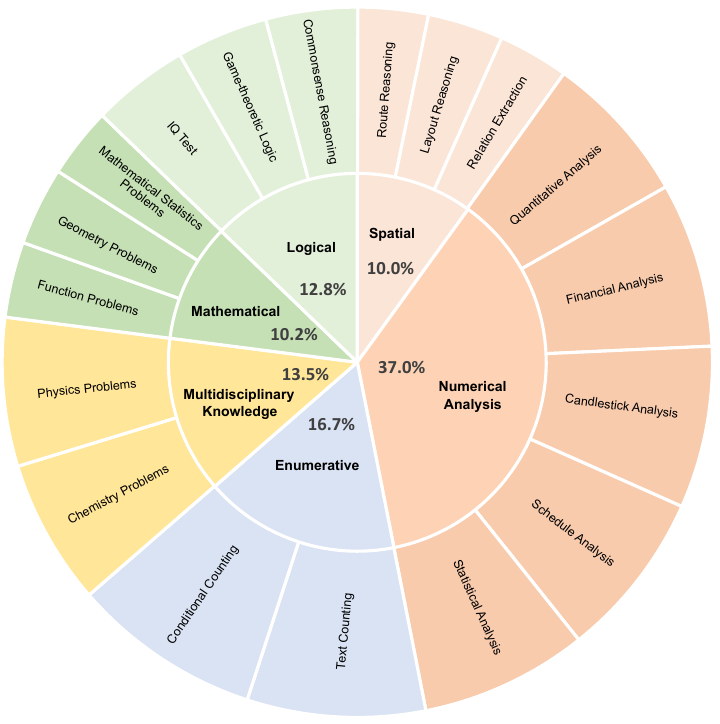}
\label{fig:distribution}
\end{minipage}
\end{figure*}

\subsection{Data Curation Framework}
\label{sec:curation}

\textbf{Dataset Collection.} We constructed the OCR-Reasoning dataset by aggregating images from three primary sources: (1) Internet-sourced images from publicly available online repositories, (2) real-world photographs capturing street views and handwritten notes, and (3) images curated from established benchmarks including InfoVQA~\citep{mathew2022infographicvqa}, DocVQA~\citep{mathew2021docvqa}, ChartQA~\citep{masry2022chartqa}, CharXiv~\citep{wang2024charxiv}, WildReceipt~\citep{sun2021spatial}, and MME-Finance~\citep{gan2024mme}. During data collection, we prioritized comprehensive coverage of text-rich scenarios commonly encountered in daily life. We also noted a severe lack of reasoning data related to handwritten content. To address this, our annotators selected and transcribed college-level problems in chemistry, physics, geometry, functions, and statistics, which were then photographed to create a set of handwritten reasoning data. In addition to college-level problems, we also include a portion of handwritten data about logical reasoning tasks. We filtered out those with low resolution or excessive noise. The final dataset comprises 1022 images, a scale comparable to previous reasoning benchmarks~\citep{lu2023mathvista,bi2025verify,xu2025visulogic}. It consists of 476 Internet-sourced images, 253 real-world photographs, and 293 images from established benchmarks. 


\textbf{Annotation Curation.}
After collecting the images, we proceed to annotate them. First, our annotators will design questions based on the images to evaluate the reasoning ability of MLLMs. To guarantee the quality of the data, we engage PhD candidates in STEM fields as expert annotators. For each image, three annotators independently propose a question. Then, other annotators score and select the highest-quality question. Subsequently, we generate reasoning processes and answers through two parallel pathways: 1. Human Annotation: Annotators manually produce one reasoning process along with the corresponding answer. 2. Model-Based Generation: We input both the questions and answers into closed-source MLLMs (e.g., Gemini 2.0 Flash) to generate an alternative reasoning process and answer.

\textbf{Data Correction.} After obtaining the questions, reasoning processes, and answers, three annotators evaluate the annotations from both pathways. The pathway with the highest average score is chosen as the final reasoning process and the corresponding answers. Finally, a manual review step is conducted to examine and correct all question-answer pairs and reasoning processes.

\textbf{Data Taxonomy.} After completing the data annotation process, we will categorize the data into six categories based on the reasoning skills required to answer the questions. To mitigate human bias, we implement a two-stage classification approach. In the initial phase, three annotators independently classified each example into one of six predefined categories. Then, we implemented a majority voting system where the final category assignment was determined by plurality consensus among the three annotators. The definitions of each category are as follows: \textbf{Spatial Reasoning} focuses on the model's ability to reason about spatial relationships between text and visual elements, as well as layout elements within text-rich images. \textbf{Numerical Analysis Reasoning} involves calculations related to numerical variations in text-rich images, including cost-effective purchase decisions, growth rate estimation, financial report analysis, schedule planning, and data interpretation. Numerical Analysis Reasoning also encompasses samples related to scenarios involving web screenshots, financial documents, or product manuals.  \textbf{Mathematical Reasoning} entails solving mathematical problems (e.g., functions, geometry, statistics) in text-rich images using mathematical knowledge. Compared to other mathematical benchmarks, the mathematical reasoning-related data in our benchmark is handwritten by our annotators, which requires models to possess stronger OCR capabilities to accomplish these tasks. \textbf{Enumerative Reasoning} focuses on counting text and visual elements in text-rich images that satisfy specific query conditions. \textbf{Logical Reasoning} requires critical thinking and drawing inferences from provided text-rich images to arrive at conclusions. \textbf{Multidisciplinary Knowledge Reasoning} involves applying cross-domain knowledge (e.g., physics, chemistry) to interpret text and visual elements in text-rich images. We provide some examples in Fig.~\ref{fig:example}.

\subsection{Dataset Statistics}
\label{sec:statistics}
The key statistics of OCR-Reasoning are summarized in Tab.~\ref{tab:statistics}. This benchmark contains 1,069 questions categorized into two distinct formats: multiple-choice (with provided answer options) and free-form responses. The free-form answers are further classified into three data types: integers, floating-point numbers, and strings. Notably, our benchmark contain extended analytical reasoning processes, evidenced by an average combined length of 421 characters for reasoning chains and final answers. The maximum length reaches 3,106 characters, highlighting the complexity of the OCR-Reasoning. As shown in Fig.~\ref{fig:distribution}, the question distribution spans six reasoning categories: Spatial Reasoning (10.0\%), Numerical Analysis (37.0\%), Logical Reasoning (12.8\%), Mathematical Reasoning (10.2\%), Multidisciplinary Knowledge (13.5\%), and Enumerative Reasoning (16.7\%). Numerical Analysis Reasoning covers 5 real-world task types, more than the 2-3 task types in other categories, hence it accounts for a larger proportion. More examples are presented in the Sec.~\ref{sec:example_of_subtasks}.

\subsection{Evaluation Protocols}
\label{sec:metric}

Following previous methods~\citep{lu2023mathvista,zhang2024mathverse}, OCR-Reasoning adopts a three-stage evaluation framework: (1) Response Generation, (2) Answer Extraction, and (3) Score Computation. First, the multimodal large language model (MLLM) processes an input query to generate detailed responses. Subsequently, an LLM-based answer extractor (e.g., GPT-4o) extracts concise answer text from these responses through semantic parsing. Our preliminary study on 200 examples shows that this extraction process achieves over 99.5\% accuracy. Finally, the extracted answers undergo normalization into standardized formats (e.g., option letters, integers, or strings) before accuracy-based metric calculation for deterministic evaluation.

For the evaluation of reasoning processes, inspired by evaluation in large language models~\citep{zheng2023judging,chang2024survey}, we employed the LLM-as-judge~\citep{zheng2023judging} approach to assess the reasoning process. Given a question, a detailed response from an MLLM, and a ground truth of the reasoning trajectory, an LLM judge is asked to directly assign a score to detailed responses. Our adoption of this methodology is based on solid empirical justification: 1) Human evaluation is costly, while LLM can quickly process large amounts of data. 2) LLM can reduce the variance among human evaluators. 3) LLM as Judge is commonly used in NLP to evaluate reasoning processes. We tried to use human-grounded validation across different models to compare their scores with those of the LLM-as-Judge: the human-grounded validation score for DouBao-1.5-Vision-Pro is 53.1 (vs. LLM-as-Judge score 55.4), for Qwen2.5-VL-72B is 50.2 (vs. LLM-as-Judge score 51.8), for Llama4-Scout-109B-A17B is 43.8 (vs. LLM-as-Judge score 44.9), and for OpenAI-o1 is 47.6 (vs. LLM-as-Judge score 48.5). The scores of human-grounded validation are close to those of the LLM-as-Judge.

\begin{table*}[t]
\centering
\caption{Accuracy scores on the OCR-Reasoning. The results include OCR + LLM, closed-source MLLMs, and open-source MLLMs. Bold denotes the best performance.}
\resizebox{\linewidth}{!}
{\begin{tabular}{l|c|c|c|c|c|c|c}
\toprule  \textbf{Method}  & \textbf{Overall} & \textbf{Spatial} & \makecell{\textbf{Numerical} \\ \textbf{Analysis}} & \textbf{Mathematical} & \textbf{Enumerative} & \textbf{Logical} & \makecell{\textbf{Multidisciplinary} \\ \textbf{Knowledge}}  \\ 
\midrule
\multicolumn{8}{c}{\textbf{OCR + LLM}} \\
\midrule
OpenAI-o3-mini~\citep{o3mini} &  \textbf{33.3} &  \textbf{17.4} &  \textbf{41.2} & \textbf{25.5} & \textbf{41.3}  & 24.3  &  27.7  \\
DeepSeek-R1-Distill-Qwen-32B~\citep{guo2025deepseek} & 26.5  & 11.9 & 28.9 & 23.5 & 34.6 & 18.8 &  30.7  \\
Qwen2.5-32B~\cite{qwen2.5} &  26.5 & 13.7 & 29.9 & 16.6 & 29.1 & \textbf{26.4} & \textbf{31.4}  \\
\midrule
\multicolumn{8}{c}{\textbf{Closed-Source MLLM}} \\
\midrule
Gemini-2.0-Flash~\citep{team2023gemini} &  39.3 & 19.3  & 47.2  & 24.5 &  49.7 & 
 36.8 & 32.1   \\
GPT-4o~\citep{hurst2024gpt} &  30.7 & 21.1 & 35.9  & 18.6 & 40.8 & 
 26.4 & 23.4   \\
OpenAI-o1~\citep{jaech2024openai} &  44.4 & \textbf{27.5}  & 46.2  & \textbf{43.1} &  50.8 & 
 \textbf{40.3} & 49.6   \\
Claude-3.7-Sonnet~\citep{anthropic2025claude37} &  35.8 & 20.2  & 35.4  & 23.5 &  \textbf{60.3} & 
 30.6 & 32.1   \\
DouBao-1.5-Vision-Pro~\citep{guo2025seed1} &  \textbf{46.8} & \textbf{27.5}  & \textbf{54.0}  & 33.3 &  50.8 & 
 34.7 & \textbf{58.4}   \\
\midrule
\multicolumn{8}{c}{\textbf{Open-Source MLLM}} \\
\midrule
Qwen2.5-VL-3B~\citep{bai2025qwen2} &  12.2 & 11.0  & 11.8  & 9.8 &  19.0 & 
 7.6 & 11.7   \\
Qwen2.5-VL-7B~\citep{bai2025qwen2} &  15.7 & 13.8  & 11.6  & 8.8 &  20.1 & 
 9.0 & 35.8  \\
Qwen2.5-VL-32B~\citep{bai2025qwen2} &  36.2 & 21.1 & 38.7 & 25.5 & 46.9 &  34.7 & 36.5  \\
Qwen2.5-VL-72B~\citep{bai2025qwen2} &  37.5 & \textbf{24.8}  & 44.7  & 22.5 &  47.5 & 28.5 & 34.3 \\
InternVL3-2B~\citep{zhu2025internvl3} &  10.8 & 11.9  & 4.8  & 7.8 &  18.4 & 
 11.8 & 18.3   \\
InternVL3-8B~\citep{zhu2025internvl3} &  11.5 & 12.8  & 5.8  & 11.8 &  17.9 & 
 7.6 & 22.6   \\
InternVL3-32B~\citep{zhu2025internvl3} & 17.1 & 14.7  & 10.3  & 14.7 &  24.0 & 
11.8 & 37.2   \\
InternVL3-78B~\citep{zhu2025internvl3} &  19.9 & 13.8  & 22.4  & 9.8 &  14.0 & 
27.1 & 25.5   \\
Llama4-Scout-109B-A17B~\citep{meta2025llama} &  27.7 & 15.6  & 34.7  & 16.7 &  41.3 & 
22.9 & 12.4   \\
Kimi-VL-A3B-Thinking~\citep{kimiteam2025kimivltechnicalreport} &  20.5 & 11.9  & 22.4  & 14.7 &  24.6 & 
21.5 & 19.7   \\
VL-Rethinker-7B~\citep{vl-rethinker} &  14.6 & 8.3  & 16.1  & 9.8 &  19.6 & 
8.3 & 19.0   \\
MM-Eureka-Qwen-7B~\citep{meng2025mm} & 13.2 & 9.2  & 7.0  & 10.8 & 18.4 & 
15.3 & 27.0   \\
VLAA-Thinker-Qwen2.5VL-7B~\citep{chen2025sftrlearlyinvestigation} & 14.4 & 11.9 & 10.3 & 7.8 & 21.2 & 11.8
 & 27.0 \\
QvQ~\citep{Qwen2-VL} &  32.7 & 24.8 & 34.7 & 15.7 & 44.1 & 31.9 & 32.1 \\
Keye-VL-8B~\citep{team2025kwai} &  22.6 & 13.8 & 21.9 & 26.5 & 25.7 & 17.4 & 30.7 \\
Thyme-RL-7B~\citep{zhang2025thyme}  & 15.2 & 12.8   & 10.8    & 10.8  & 20.7 & 18.1   & 23.4 \\
DeepEyesV2~\citep{hong2025deepeyesv2} & 20.9 & 11.9 & 18.8 & 13.7 & 27.9 & 18.85 & 32.8 \\
MiMo-VL-RL-7B~\citep{coreteam2025mimovltechnicalreport} & 38.8 & 20.2 & 41.2 & 22.5 & 51.4 & \textbf{38.9} & 42.3 \\
GLM-4.1V-Thinking-9B~\citep{hong2025glm} & \textbf{44.1} & 22.9 & \textbf{49.2} & \textbf{35.3} & \textbf{53.1} & 35.4 & \textbf{50.4} \\
\midrule
\multicolumn{8}{c}{\textbf{Document-Oriented MLLMs}} \\
\midrule
mPLUG-DocOwl2-8B~\citep{hu2024mplug} & 3.3 & 3.7 & 0.3 & 1.0 & 7.3 & 9.7 & 1.5 \\
Docopilot-8B~\citep{duan2025docopilot} & 11.6 & 11.9 & 6.5 & 6.9 & 19.6 & 8.3 & 22.6 \\
DocMark-2B~\citep{xiao2025adaptive} & 7.4 & 8.3 & 0.3 & 6.9 & 14.5 & 3.5 & 22.6 \\
TokenVL-8B~\citep{guan2025token} & 14.3 & 10.1 & 8.8 & 8.9 & 25.7 & 13.9 & 23.4 \\

\bottomrule
\end{tabular}}
\label{tab:accuracy_metric}
\end{table*}

\section{Experiment}
\label{sec:experiment}
In this section, we conduct a comprehensive evaluation of existing MLLMs on OCR-Reasoning. We first describe the experimental setup in Sec.~\ref{sec:setup}. Then, the overall results and the corresponding analysis are presented in Sec.~\ref{sec:results}. 

\subsection{Experiment Setup}
\label{sec:setup}

\textbf{Evaluation Models.} We evaluate three distinct types of foundation models on OCR-Reasoning: (a) Large Language Models (LLMs) with OCR results (Extracting by PP-OCRv3~\citep{li2022pp}), including Deepseek-R1~\citep{zhou2025r1} and OpenAI-o3-mini. (b) closed-source MLLMs, comprising Gemini-2.0-Flash~\citep{deepmind2025flashthinking}, GPT-4o~\citep{hurst2024gpt}, OpenAI-o1~\citep{jaech2024openai}, Claude-3.7-Sonnet~\citep{anthropic2025claude37}, and DouBao-1.5-Vision-Pro~\citep{guo2025seed1}. (c) Open-source MLLMs, represented by models like Qwen2.5-VL~\citep{bai2025qwen2}, InternVL3~\citep{zhu2025internvl3}, Llama4-Scout~\citep{meta2025llama}, Kimi-VL~\citep{kimiteam2025kimivltechnicalreport}, VL-Rethinker~\citep{vl-rethinker}, MM-Eureka~\citep{meng2025mm}, VLAA-Thinker~\citep{chen2025sftrlearlyinvestigation}. (d) Document-Oriented MLLMs, including mPLUG-Docow2~\citep{hu2024mplug25}, Docopilot~\citep{duan2025docopilot}, DocMark~\citep{xiao2025adaptive}, and TokenVL~\citep{guan2025token}.

\textbf{Implementation Details.} To evaluate the generalization capacity of Multimodal Large Language Models (MLLMs), we adopt a zero-shot evaluation protocol without model fine-tuning or few-shot prompting. Following the standardized chain-of-thought paradigm, we present MLLMs with both visual inputs (images) and textual questions, accompanied by explicit instructions: ``Solve the complex problem through step-by-step reasoning." For text-only Large Language Models (LLMs), we substitute visual inputs with the OCR results (using PP-OCRv3~\citep{li2022pp} to obtain the OCR results) while retaining identical textual queries. Given the inherent variability in output formats across text-rich image scenarios (e.g., monetary values like \$15, temporal expressions like 20 days, or timestamps like 19:00:00), we implement format-specific prompting. This involves appending the directive: ``The composition of the final answer should be: xxxxx" to each query. For instance, when expecting currency outputs ``\$15'', the format-specific prompting is: ``The composition of the final answer should be: \$ + Integer". 

\begin{table*}[t]
\centering
\caption{Impact of Chain-of-Thought prompting on different MLLMs.}
\resizebox{\linewidth}{!}
{\begin{tabular}{l|c|c|c|c|c|c|c|c}
\toprule   \textbf{Method} & \textbf{CoT} & \textbf{Overall} & \textbf{Spatial} & \makecell{\textbf{Numerical} \\ \textbf{Analysis}} & \textbf{Mathematical} & \textbf{Enumerative} & \textbf{Logical} & \makecell{\textbf{Multidisciplinary} \\ \textbf{Knowledge}}  \\ 
\midrule
Qwen2.5-VL-32B~\citep{bai2025qwen2} & $\times$ &  33.0 & 12.8 & 33.7 & 24.5 & 48.0 & 
 28.4 & 38.7  \\
Qwen2.5-VL-32B~\citep{bai2025qwen2} & $\checkmark$ &  36.2 & 21.1 & 38.7 & 25.5 & 46.9 & 
 34.7 & 36.5  \\
GPT-4o~\citep{hurst2024gpt}  & $\times$ &  26.5 & 11.9 & 33.4 & 15.7 & 29.1 & 25.0 & 24.1   \\
GPT-4o~\citep{hurst2024gpt} & $\checkmark$  &  30.7 & 21.1 & 35.9  & 18.6 & 40.8 & 
 26.4 & 23.4   \\
Kimi-VL-A3B-Thinking~\citep{kimiteam2025kimivltechnicalreport} & $\times$ & 20.1 & 11.0 & 19.1 & 16.7 & 30.2 & 
19.4 & 20.4   \\
Kimi-VL-A3B-Thinking~\citep{kimiteam2025kimivltechnicalreport} & $\checkmark$ &  20.5 & 11.9  & 22.4  & 14.7 &  24.6 & 
21.5 & 19.7   \\
VL-Rethinker-7B~\citep{vl-rethinker} & $\times$ &  19.1 & 13.7  & 16.6  & 9.8 &  25.7 & 
14.6 & 33.6   \\
VL-Rethinker-7B~\citep{vl-rethinker} & $\checkmark$ &  14.6 & 8.3  & 16.1  & 9.8 &  19.6 & 
8.3 & 19.0   \\
MM-Eureka-Qwen-7B~\citep{meng2025mm} & $\times$ &  12.2 & 10.1 & 6.3 & 8.8 & 16.8 & 
14.6 & 25.5   \\
MM-Eureka-Qwen-7B~\citep{meng2025mm} & $\checkmark$  & 13.2 & 9.2  & 7.0  & 10.8 & 18.4 & 
15.3 & 27.0   \\
\bottomrule
\end{tabular}}
\label{tab:cot}
\end{table*}

\subsection{Overall Results}
\label{sec:results}

\textbf{The use of visual images as input is crucial.} To assess its importance, we replace the images with their OCR results and feed them into the LLMs for comparison. As shown in Tab.~\ref{tab:accuracy_metric}, substituting image input with OCR text leads to a significant decline in model performance. For instance, when using the same LLM, the performance of even a strong reasoning model like DeepSeek-R1-Distill-Qwen-32B remains 9.7\% lower than that of Qwen2.5-VL-32B. This demonstrates the critical importance of image input for text-rich image reasoning tasks. We present some qualitative results in Appendix~\ref{sec:qualitative_ocr}.

\textbf{The performance of current MLLMs still has significant room for improvement.} Specifically, our analysis reveals several key observations: 1. As shown in Tab.~\ref{tab:accuracy_metric}, the top-performing model is Doubao-1.5-Vision-Pro. While Doubao-1.5-Vision-Pro performs strongly on text-rich image understanding tasks—such as DocVQA (96.7\%), InfoVQA (89.3\%), and ChartQA (87.4\%)—its performance on OCR-Reasoning does not exceed 50\%. This highlights the particular challenge of integrating visual, textual, and logical information in reasoning scenarios. 2. Among different reasoning types, MLLMs perform most strongly on enumerative reasoning, which consistently ranks as the first or second best capability in both closed-source and open-source models. 3. Furthermore, scaling up model parameters is positively correlated with performance gains, as illustrated by the Qwen2.5-VL series: the 7B model surpasses the 3B version by 3.5\%, and the 32B model outperforms the 7B version by 20.5\%. 4. Additionally, we observe that document-oriented MLLMs still face difficulties in complex reasoning. Although document-oriented MLLMs are effective at basic comprehension, their limitations in deeper reasoning underscore the need for innovations in model architecture or training strategies.

\begin{table*}[t]
\centering
\caption{Reasoning scores on the OCR-Reasoning benchmark. Bold denotes the best performance.}
\resizebox{\linewidth}{!}
{\begin{tabular}{l|c|c|c|c|c|c|c}
\toprule   \textbf{Method}  & \textbf{Overall} & \textbf{Spatial} & \makecell{\textbf{Numerical} \\ \textbf{Analysis}} & \textbf{Mathematical} & \textbf{Enumerative} & \textbf{Logical} & \makecell{\textbf{Multidisciplinary} \\ \textbf{Knowledge}}  \\ 
\midrule
\multicolumn{8}{c}{\textbf{Closed-Source MLLM}} \\
\midrule
Gemini-2.0-Flash~\citep{team2023gemini} &  49.5 & 31.5 & 57.1 & 42.6 & 49.3 & 
 47.4 & 49.2   \\
GPT-4o~\citep{hurst2024gpt} &  45.4 & 35.4  & 48.9  & 33.0 & 48.7 & 48.0 & 45.5   \\
OpenAI-o1~\citep{jaech2024openai} &  48.5 & 36.9  & 53.9  & 50.0 & 39.4 & 
 49.4 & 51.8   \\
Claude-3.7-Sonnet~\citep{anthropic2025claude37} & 50.3 & 37.7 & 55.0  & 38.8 & \textbf{58.1} &  48.6 & 46.5   \\
DouBao-1.5-Vision-Pro~\citep{guo2025seed1} & \textbf{55.4} & 38.2  & \textbf{61.8}  & \textbf{50.2} & 52.4 & 
 52.8 & \textbf{61.2}   \\
 \midrule
\multicolumn{8}{c}{\textbf{Open-Source MLLM}} \\
\midrule
Qwen2.5-VL-3B~\citep{bai2025qwen2} & 22.3 & 18.5 & 25.6 & 15.7 & 22.9 & 20.8 & 21.3 \\
Qwen2.5-VL-7B~\citep{bai2025qwen2} & 34.0 & 24.9 & 39.2 & 27.5 & 41.5 & 28.9 & 27.3 \\
Qwen2.5-VL-32B~\citep{bai2025qwen2} & 54.6 & \textbf{38.5} & 58.9 & 45.9 & 55.8 & \textbf{56.3} & 57.9 \\
Qwen2.5-VL-72B~\citep{bai2025qwen2} & 51.8 & 35.9 & 58.6 & 41.3 & 54.7 & 49.9 & 50.8 \\
InternVL3-2B~\citep{zhu2025internvl3} & 15.7 & 15.3 & 13.4 & 13.3 & 21.4 & 16.5 & 15.7 \\
InternVL3-8B~\citep{zhu2025internvl3} & 16.3 & 16.1 & 14.9 & 13.8 & 21.5 & 15.0 & 16.6 \\
InternVL3-32B~\citep{zhu2025internvl3} & 42.6 & 32.1 & 43.9 & 38.0 & 45.3 & 43.6 & 46.1 \\
InternVL3-78B~\citep{zhu2025internvl3} & 43.3 & 29.2 & 50.3 & 35.5 & 38.6 & 46.9 & 42.2 \\
Llama4-Scout-109B-A17B~\citep{meta2025llama} & 44.9 & 33.0 & 49.3 & 36.7 & 47.1 & 45.4 & 44.1 \\
Kimi-VL-A3B-Thinking~\citep{kimiteam2025kimivltechnicalreport} & 40.8 & 30.9 & 43.1 & 37.4 & 45.6 & 38.5 & 40.7 \\
VL-Rethinker-7B~\citep{vl-rethinker} & 29.8 & 23.1 & 33.7 & 23.7 & 31.4 & 26.5 & 29.8 \\
MM-Eureka-Qwen-7B~\citep{meng2025mm} & 21.9 & 20.7 & 20.5 & 16.5 & 24.8 & 22.2 & 26.9 \\
VLAA-Thinker-Qwen2.5VL-7B~\citep{chen2025sftrlearlyinvestigation} & 24.8 & 22.2 & 25.8 & 18.0 & 25.9 & 24.2 & 28.5 \\
\midrule
\multicolumn{8}{c}{\textbf{Document-Oriented MLLMs}} \\
\midrule
mPLUG-DocOwl2-8B~\citep{hu2024mplug} & 12.9 & 13.6 & 12.6 & 10.3 & 13.8 & 15.6 & 10.9 \\
Docopilot-8B~\citep{duan2025docopilot} & 20.6 & 17.1 & 22.4 & 13.2 & 25.2 & 19.6 & 19.1 \\
DocMark-2B~\citep{xiao2025adaptive} & 15.3 & 13.9 & 13.5 & 12.3 & 22.7 & 15.3 & 13.9 \\
\bottomrule
\end{tabular}}
\label{tab:reasoning_metric}
\end{table*}



\textbf{CoT prompting performs differently across different models.} The results of the influence of CoT prompts are presented in Tab.~\ref{tab:cot}. We observe that CoT prompting performs differently across different models. On most models, CoT prompting consistently enhances their capabilities. This improvement is particularly pronounced in spatial reasoning, where CoT prompting exerts the most significant impact. Specifically, CoT prompting improves performance by 3.2\% on Qwen2.5-VL-32B, and 4.2\% on GPT-4o, respectively. However, for VL-Rethinker-7B~\citep{vl-rethinker}, the application of CoT prompting typically results in performance degradation. This phenomenon may stem from the forced rethinking machine on VL-Rethinker-7B. Adding an additional CoT prompt during inference creates a discrepancy between training and testing conditions, ultimately leading to reduced performance.

\textbf{Impact of Reasoning path.}
The scores of the reasoning path are presented in Tab.~\ref{tab:reasoning_metric}. Overall, we observe that the ranking based on Reasoning Path scores aligns with that based on final answer accuracy, with the exception of Gemini and Claude-3.7-Sonnet. Specifically, the high scores of Gemini and Claude-3.7-Sonnet are primarily due to the high quality of their reasoning path. We find that for many erroneous samples, the reasoning steps are largely sound, with only minor errors leading to the incorrect outcome. Additional qualitative analyses are provided in the appendix (Sec.~\ref{sec:qualitative_reasoning}) to further illustrate these observations.

\textbf{Some reinforcement learning methods perform poorly on text-rich image reasoning tasks.} The performance of some reinforcement learning methods on text-rich image reasoning tasks is relatively poor compared to their baseline. There are several possible reasons for this. First, the reward function: The reward functions in these reinforcement learning methods are not specifically designed for text-rich image reasoning tasks. Most existing reward functions are tailored for mathematical reasoning tasks. How to design a reward function applicable to text-rich image reasoning tasks is a highly worthwhile research direction. Second, a notable discrepancy exists between the training data and the benchmark. The majority of training data is primarily designed for printed mathematical problems, while our benchmark contains data from a wide variety of scenarios. How to select training data to improve OCR inference performance is a highly valuable research direction.

\textbf{Impact of Few-shot Prompting.} We also conducted experiments to explore the impact of few-shot prompting. We annotated three additional samples as few-shot demonstrations and validated the performance of one-shot and three-shot prompting on Qwen2.5-VL-7B. As shown in the Tab.~\ref{tab:few_shot}, few-shot prompting improves overall performance—particularly on subtasks requiring adherence to specific logical steps (e.g., Numerical Analysis Reasoning and Logical Reasoning). This demonstrates that moderate task-specific guidance helps the model understand and comply with task requirements. We observed a decline in the performance of Multidisciplinary Knowledge Reasoning. The potential reasons may be: The increased length of input tokens caused by few-shot examples, combined with the extended reasoning path inherently required by multidisciplinary knowledge reasoning, poses a significant challenge to the model's long-text processing and reasoning capabilities.

\begin{table*}[t]
\centering
\caption{Impact of few-shot prompting on Qwen2.5-VL-7B.}
\resizebox{\linewidth}{!}
{\begin{tabular}{l|c|c|c|c|c|c|c}
\toprule   \textbf{Method} & \textbf{Overall} & \textbf{Spatial} & \makecell{\textbf{Numerical} \\ \textbf{Analysis}} & \textbf{Mathematical} & \textbf{Enumerative} & \textbf{Logical} & \makecell{\textbf{Multidisciplinary} \\ \textbf{Knowledge}}  \\ 
\midrule
Qwen2.5-VL-7B~\citep{bai2025qwen2} &  15.7 & 13.8  & 11.6  & 8.8 &  20.1 & 
 9.0 & 35.8  \\
One-shot prompting &  16.1 & 12.8 & 14.6 & 10.8 & 22.3 & 13.9 & 21.1   \\
Three-shot prompting &  16.4 & 13.7 & 14.8  & 10.0 & 22.9 & 13.2 & 22.3  \\
\bottomrule
\end{tabular}}
\label{tab:few_shot}
\end{table*}

\textbf{The ``thinking with images" approach represents a promising direction for enhancing reasoning capabilities over text-rich images.} ``Thinking with images" has demonstrated tremendous potential in general-scene reasoning; therefore, we also evaluated the two latest "thinking with images" methods on the OCR-Reasoning task. The experimental results are presented in Table~\ref{tab:accuracy_metric}, which illustrate the potential of this approach to enhance the model's reasoning ability on text-rich images.

\subsection{Comparison with Existing Benchmarks}
We conducted experiments to quantitatively compare our OCR-Reasoning dataset with existing benchmarks (including DocVQA, ChartQA, TextVQA, OCRBench) using three multimodal large language models (MLLMs): Qwen2.5-VL-7B, InternVL3-8B, and TokenVL-8B. The experimental results are presented in Tab.~\ref{tab:benchmark_comparison}. As observed, while existing methods achieve strong performance on conventional datasets, they exhibit significant performance degradation on text-rich image reasoning tasks. This discrepancy stems from the fundamental difference in evaluation focus: existing datasets primarily assess models’ perceptual capabilities (e.g., text detection, recognition, and basic visual-linguistic alignment), whereas OCR-Reasoning requires the model to achieve accurate perception and further conduct thinking and reasoning. These findings indicate that the text-rich image reasoning ability of current MLLMs still has substantial room for improvement.

\begin{table}[htbp]
  \centering
  \caption{Comparison with Existing Benchmarks}
  \resizebox{0.7\linewidth}{!}
{
    \begin{tabular}{l|c|c|c|c|c}
    \toprule
    Models & DocVQA & ChartQA & OCRBench & TextVQA & OCR-Reasoning \\
    \midrule
    Qwen2.5-VL-7B & 95.7   & 87.3   & 864  & 84.9  & 15.7 \\
    InternVL3-8B & 92.7   & 86.6   & 880  & 80.2   & 11.5 \\
    TokenVL-8B & 94.2   & 86.6   & 860  & 79.9  & 14.3 \\
    \bottomrule
    \end{tabular}}%
  \label{tab:benchmark_comparison}%
\end{table}%

\subsection{Error Analysis}
\label{sec:erroranalysis}

\begin{wrapfigure}{r}{0.45\textwidth}
  \vspace{-5ex}
  \begin{center}
    \includegraphics[width=0.45\textwidth]{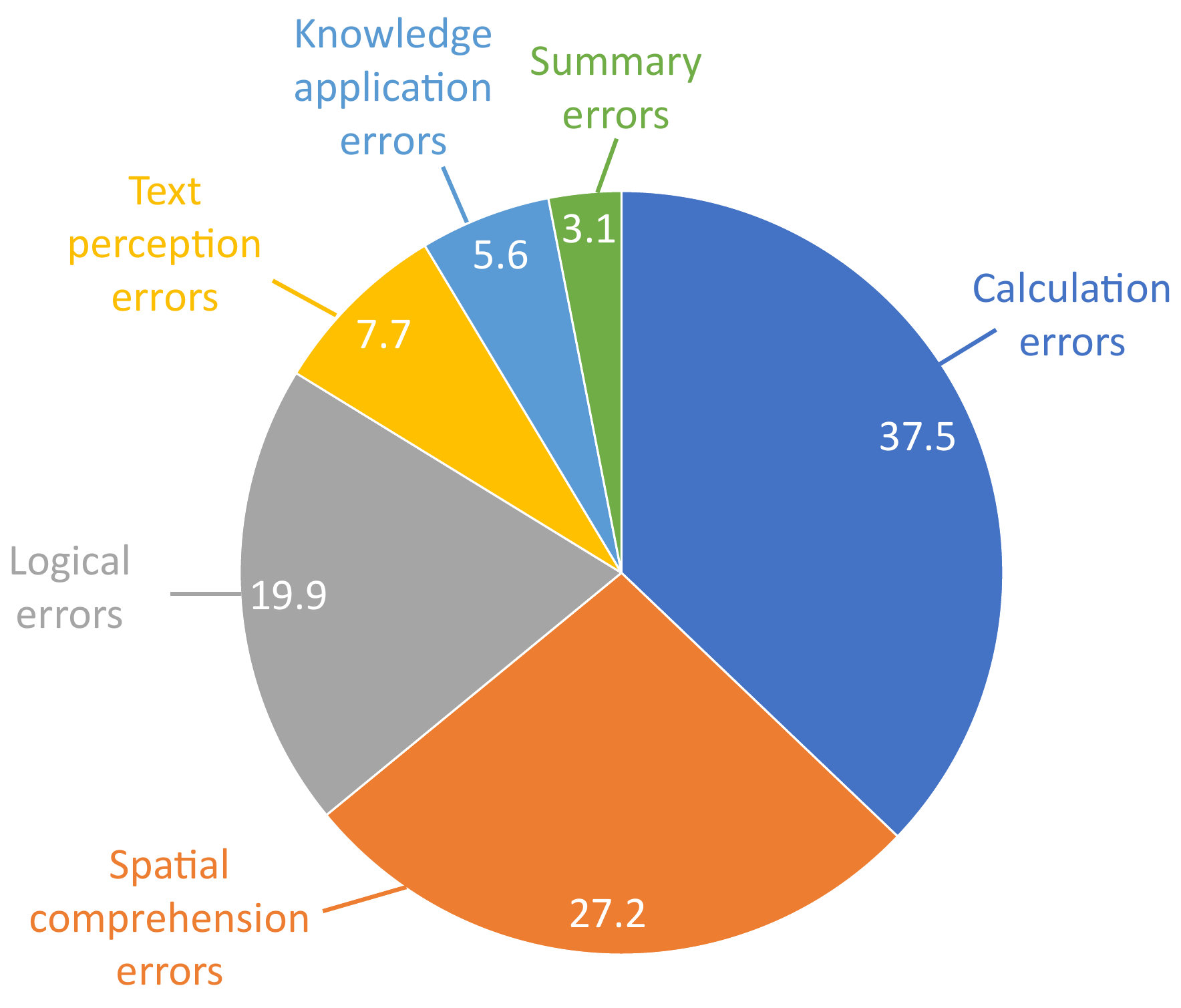}
  \end{center}
  \caption{
  Error analysis for DouBao-1.5-Vision-Pro reveals six main issues: Calculation errors, Spatial comprehension errors, Logical errors, Text perception errors, Knowledge application errors, and Summary errors.
  }
  \vspace{-2mm}
  \label{fig:error_analysis}
\end{wrapfigure}

We analyzed the error types of DouBao-1.5-Vision-Pro and classified them into six major categories: (1) calculation errors (37.5\%), (2) spatial comprehension errors (27.2\%), (3) logical errors (19.9\%), (4) text perception errors (7.7\%), (5) knowledge application errors (5.6\%), and (6) summary errors (3.1\%). Calculation errors occur when layout understanding and text perception are correct, but mistakes are made during the calculation process. Spatial comprehension errors arise when the model fails to properly understand the spatial layout information of text in images. Logical errors result from unreasonable assumptions or inverted cause-and-effect relationships. Knowledge application errors occur in cases such as the incorrect application of theorems or misunderstanding of common sense. Text perception errors happen when text recognition is incorrect. Summary errors occur when the reasoning process is normal but the final answer is incorrect. This distribution reveals that the DouBao-1.5-Vision-Pro's primary challenges lie in higher-level cognitive tasks rather than basic perception tasks.

\section{Conclusion}
\label{sec:conclusion}
In this paper, we introduce OCR-Reasoning, a comprehensive benchmark designed to systematically evaluate the reasoning capabilities of state-of-the-art Multimodal Large Language Models (MLLMs) in text-rich image scenarios. The benchmark provides a structured framework comprising 1,069 human-annotated examples, which cover 6 core reasoning abilities across 18 distinct tasks. In contrast to existing benchmarks that focus solely on final-answer accuracy, OCR-Reasoning incorporates annotations for both the final answers and the corresponding step-by-step reasoning processes. This dual annotation enables a more comprehensive evaluation of model reasoning. Through extensive experiments, we reveal critical limitations of current MLLMs and identify potential avenues for future improvement.

\noindent\textbf{Limitations.} This work has two main limitations. First, since most of our data collection and annotation processes are performed manually, the high costs associated with these processes have resulted in our dataset size being comparable to previous methods~\citep{liu2024ocrbench,xu2025visulogic}. In the future, we plan to combine automated annotation with human efforts to expand the dataset scale. Second, following the evaluation process of LLMs, we employed the LLMs-as-Judges approach to assess model reasoning processes. However, issues such as biases in LLMs-as-Judges, adversarial attacks, and inherent weaknesses in the methodology may affect evaluation accuracy~\citep{li2024llmsasjudgescomprehensivesurveyllmbased}. We intend to develop more robust evaluation approaches in future work.

\subsubsection*{Acknowledgments}
This research is supported in part by the National Natural Science Foundation of China (Grant No.:62476093).

\bibliography{iclr2026_conference}
\bibliographystyle{iclr2026_conference}

\appendix
\section{Appendix}

\subsection{The Use of Large Language Models(LLMs)}
We used LLMs only to polish and correct the grammar of this paper. This involved rephrasing sentences, checking grammar, and improving the overall flow for better readability and clarity. All research ideation, experiments, and analyses were developed by the authors, with no involvement from the LLM in these stages.

\subsection{Qualitative Analysis of Reasoning Path}
\label{sec:qualitative_reasoning}
Here, we illustrate this with an example. As shown in Fig.~\ref{fig:reasoning_analysis}(a), (b), and(c), we observe that in models such as Gemini and Claude-3.7-Sonnet, many errors occur where minor mistakes in the reasoning process lead to incorrect final answers, while the overall reasoning process remains fundamentally sound. Consequently, the LLM assigns a relatively high score to their problem-solving procedures.

\subsection{Qualitative Analysis of OCR Input}
\label{sec:qualitative_ocr}
We have provided some examples where OCR errors led to incorrect final results, as shown in Fig.~\ref{fig:reasoning_analysis}(d). We found that LLMs generally struggle to provide correct answers when OCR results contain errors.

\subsection{Qualitative Analysis of Error Case}
\label{sec:qualitative_error}

\begin{figure}[t!]
\centering
    \includegraphics[width=0.7\linewidth]{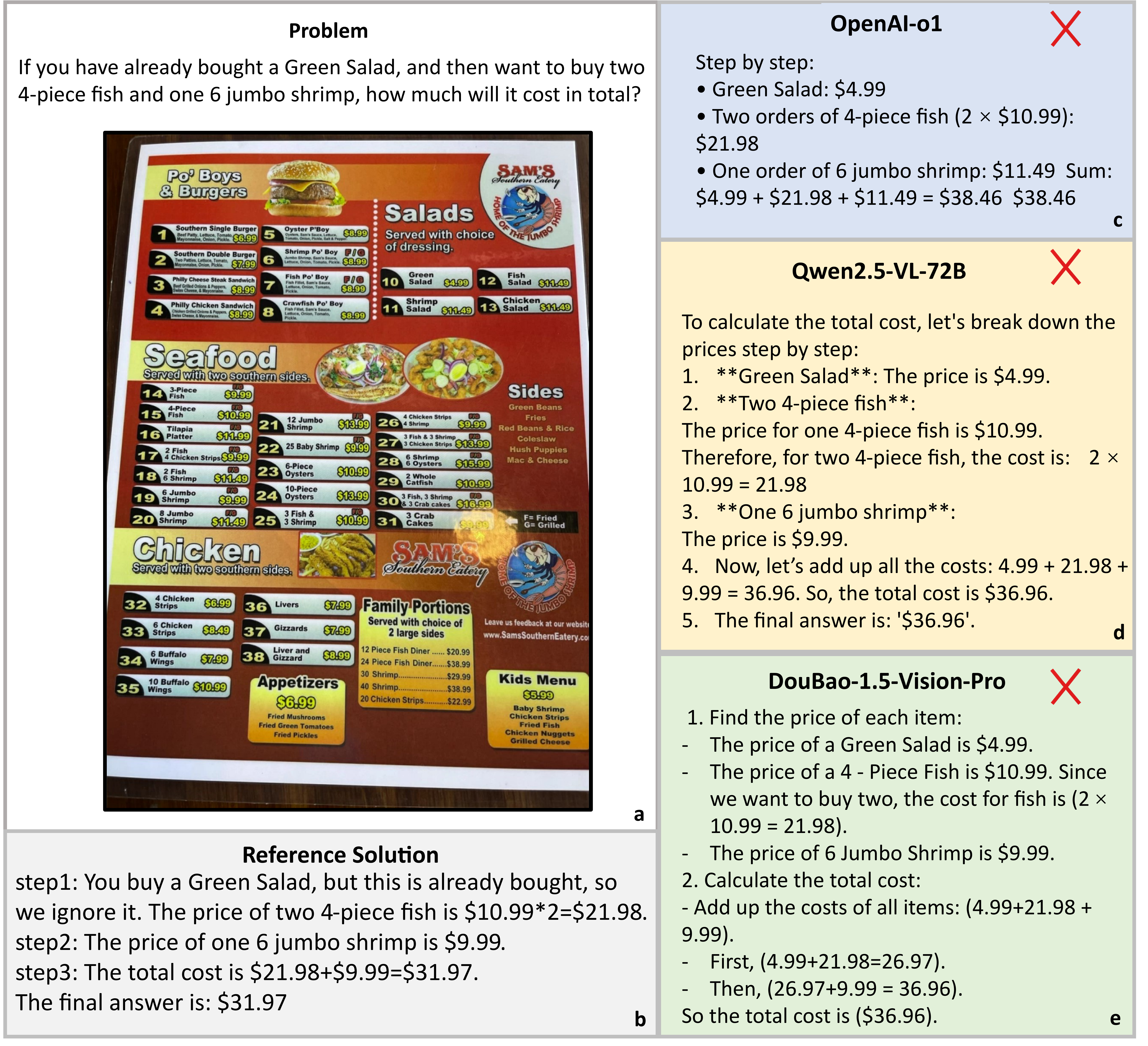}
    \caption{
        Solution examples generated by different models. (a) presents the input image and corresponding question; (b) shows the reference solution; (c)-(e) demonstrate outputs generated by different AI systems: (c) OpenAI-o1 model, (d) Qwen2.5-VL-72B model, and (e) DouBao-1.5-Vision-Pro model.
    }
    \label{fig:error}
\end{figure}

Fig.~\ref{fig:error} presents a systematic comparison of MLLM reasoning processes. For OpenAI-o1, the main error lies in its failure to correctly understand the problem. In the scenario, the Green Salad had already been purchased and paid for, but OpenAI-o1 still included it when calculating the new price. As for DouBao-1.5-Vision-Pro, it exhibits similar issues to O1 and additionally made a mistake in understanding the product name. According to the menu, ``6 jumbo shrimp'' refers to a single dish instead of six individual jumbo shrimp. However, both Qwen2.5-VL-72B and DouBao-1.5-Vision-Pro misinterpret it as six separate food items.  
made the same error as DouBao-1.5-Vision-Pro.

\subsection{Use the Data Curation Framework to Annotate Training Data}
\label{sec:training_annotation}
We leveraged our annotation logic to replace human annotators with DouBao-1.5-Vision-Pro, selecting 5000 images from DocVQA, InfoVQA, TextVQA, and ChartQA for synthetic data generation. As presented in Tab.~\ref{tab:synthetic_training}, the results of fine-tuning Qwen2.5-VL-7B using this synthetic data. These results demonstrate that the proposed data curation framework cannot only be used for building benchmark data but also be migrated to the construction of training datasets.

\begin{table*}[t]
\centering
\caption{Performance comparison of Qwen2.5-VL-7B before and after fine-tuning on the synthetic data generated by our data curation framework.}
\resizebox{\linewidth}{!}
{
    \begin{tabular}{lccccccc} 
        \toprule   
        \textbf{Method}  & \textbf{Overall} & \textbf{Spatial} & \makecell{\textbf{Numerical} \\ \textbf{Analysis}} & \textbf{Mathematical} & \textbf{Enumerative} & \textbf{Logical} & \makecell{\textbf{Multidisciplinary} \\ \textbf{Knowledge}}  \\ 
        \midrule
        Qwen2.5-VL-7B (base model) & 15.7  & 13.8  & 11.6 & 8.8 & 20.1 & 9.0 & 35.8 \\
        Qwen2.5-VL-7B (Fine-tune) & 17.0  & 14.7 & 10.6 & 21.6 & 22.9 & 17.4 & 26.3 \\        
        \bottomrule
    \end{tabular}
}
\label{tab:synthetic_training}
\end{table*}

\subsection{A Roadmap for OCR-Reasoning}
\label{sec:roadmap}
We outline a concrete roadmap for how OCR-Reasoning might evolve into a standardized benchmark suite for multimodal reasoning. We outline three core steps:

1. For reward design, we will leverage our annotation logic, replace the manual parts with leading models, and construct a batch of training data through standardized processes. Using this data, we will train a process-aware reward model that evaluates not only answer correctness but also the rationality of reasoning steps—delivering more granular, objective signals than rule-based reward functions. The model will be fully open-sourced to help the community systematically enhance models’ reasoning capabilities on text-rich images.

2. We will make the benchmark open-source on Hugging Face to facilitate downloading and use by researchers in the community. Subsequently, we will open-source the code on commonly used evaluation frameworks for MLLMs, enabling researchers to quickly evaluate their models' performance.

3. Existing benchmarks often focus on evaluating capabilities in a single dimension. This benchmark (OCR-Reasoning) focuses on assessing models' reasoning abilities on single images. Therefore, for dataset expansion, we will include more complex tasks such as reasoning on multilingual text, domain-specific documents, or multi-page handwritten materials.

\subsection{The Underlying Mechanisms of VL-Rethinker-7B}
\label{sec:rethinker}
When testing RL-trained models, we used their official prompts to ensure consistency between training and testing. On this basis, we further added the CoT prompt: 'Solve the complex problem through step-by-step reasoning.' We found that this approach improved performance on MM-Eureka, but degraded performance on VL-Rethinker. We hypothesize that this is due to a conflict between the generic CoT prompt and the model's built-in reflection/reasoning mechanism.

To further validate this, we conducted an ablation study by removing the model's official built-in prompt and then testing the effect of the generic CoT prompt. The results are presented below. The performance improved when the official prompt was removed (comparing the third and fourth rows). It supports our hypothesis that the degradation occurs specifically because the two prompts conflict within the model's input processing.

\begin{table*}[thbp]
  \centering
  \caption{Ablation study of the underlying mechanisms of VL-Rethinker-7B.}
  \label{tab:ablation_cot_prompt} 
  \resizebox{\linewidth}{!}{
    \begin{tabular}{l|c|c|c|c|c|c|c|c|c} 
      \toprule  
      Method               & Official Prompt & CoT  & Overall & Spatial & Numerical Analysis & Mathematical & Enumerative & Logical & Multidisciplinary Knowledge \\
      \midrule 
      Official Setting     & \checkmark  & $\times$ & 19.1    & 13.7    & 16.6               & 9.8          & 25.7        & 14.6    & 33.6                        \\
      Original + CoT       & \checkmark  & \checkmark & 14.6    & 8.3     & 16.1               & 9.8          & 19.6        & 8.3     & 19.0                        \\
      Ablation 1 (No Prompt)    & $\times$    & $\times$ & 17.5    & 13.8    & 15.8               & 7.8          & 21.2        & 16.7    & 28.5                        \\
      Ablation 2 (CoT)     & $\times$   & \checkmark & 18.7    & 14.7    & 13.1               & 11.7         & 27.9        & 18.8    & 31.3                        \\
      \bottomrule 
    \end{tabular}}
\end{table*}

\subsection{Practical Reasoning Tasks}
\label{sec:example_of_subtasks}

In this Section, we provide some examples of the 18 practical reasoning tasks in text-rich visual scenarios, as presented in Fig.~\ref{fig:example_of_subtasks}.

\begin{figure}[h!]
\centering
    \includegraphics[width=\linewidth]{figs/example_of_subtasks.pdf}
    \caption{
        Examples of the 18 practical reasoning tasks in text-rich visual scenarios.
    }
    \label{fig:example_of_subtasks}
\end{figure}

\begin{figure}[t!]
\centering
    \includegraphics[width=\linewidth]{figs/reasoning_analysis.pdf}
    \caption{
        Qualitative analysis of the reasoning path. (a) presents the input image and corresponding question; (b) shows the reference solution serving as ground truth; (c) demonstrates outputs generated by Gemini-2.0-Flash. (d) demonstrates outputs generated by O3-mini with OCR results input.
    }
    \label{fig:reasoning_analysis}
\end{figure}

\end{document}

%% file: math_commands.tex
\usepackage{amsmath,amsfonts,bm}

\def\eqref#1{equation~\ref{#1}}

\def\1{\bm{1}}

\DeclareMathAlphabet{\mathsfit}{\encodingdefault}{\sfdefault}{m}{sl}
\SetMathAlphabet{\mathsfit}{bold}{\encodingdefault}{\sfdefault}{bx}{n}



%% file: iclr2026_conference.bib
@inproceedings{hudson2019gqa,
  title={Gqa: A new dataset for real-world visual reasoning and compositional question answering},
  author={Hudson, Drew A and Manning, Christopher D},
  booktitle={Proceedings of the IEEE/CVF conference on computer vision and pattern recognition},
  pages={6700--6709},
  year={2019}
}

@article{lu2022learn,
  title={Learn to explain: Multimodal reasoning via thought chains for science question answering},
  author={Lu, Pan and Mishra, Swaroop and Xia, Tanglin and Qiu, Liang and Chang, Kai-Wei and Zhu, Song-Chun and Tafjord, Oyvind and Clark, Peter and Kalyan, Ashwin},
  journal={Advances in Neural Information Processing Systems},
  volume={35},
  pages={2507--2521},
  year={2022}
}

@inproceedings{mathew2021docvqa,
  title={Docvqa: A dataset for vqa on document images},
  author={Mathew, Minesh and Karatzas, Dimosthenis and Jawahar, CV},
  booktitle={Proceedings of the IEEE/CVF winter conference on applications of computer vision},
  pages={2200--2209},
  year={2021}
}

@inproceedings{masry2022chartqa,
  title={ChartQA: A Benchmark for Question Answering about Charts with Visual and Logical Reasoning},
  author={Masry, Ahmed and Do, Xuan Long and Tan, Jia Qing and Joty, Shafiq and Hoque, Enamul},
  booktitle={Findings of the Association for Computational Linguistics: ACL 2022},
  pages={2263--2279},
  year={2022}
}

@inproceedings{mathew2022infographicvqa,
  title={Infographicvqa},
  author={Mathew, Minesh and Bagal, Viraj and Tito, Rub{\`e}n and Karatzas, Dimosthenis and Valveny, Ernest and Jawahar, CV},
  booktitle={Proceedings of the IEEE/CVF Winter Conference on Applications of Computer Vision},
  pages={1697--1706},
  year={2022}
}

@inproceedings{ye2023ureader,
  title={UReader: Universal OCR-free Visually-situated Language Understanding with Multimodal Large Language Model},
  author={Ye, Jiabo and Hu, Anwen and Xu, Haiyang and Ye, Qinghao and Yan, Ming and Xu, Guohai and Li, Chenliang and Tian, Junfeng and Qian, Qi and Zhang, Ji and others},
  booktitle={The 2023 Conference on Empirical Methods in Natural Language Processing},
  year={2023}
}

@article{wei2022chain,
  title={Chain-of-thought prompting elicits reasoning in large language models},
  author={Wei, Jason and Wang, Xuezhi and Schuurmans, Dale and Bosma, Maarten and Xia, Fei and Chi, Ed and Le, Quoc V and Zhou, Denny and others},
  journal={Advances in Neural Information Processing Systems},
  volume={35},
  pages={24824--24837},
  year={2022}
}

@article{liu2024textmonkey,
  title={Textmonkey: An ocr-free large multimodal model for understanding document},
  author={Liu, Yuliang and Yang, Biao and Liu, Qiang and Li, Zhang and Ma, Zhiyin and Zhang, Shuo and Bai, Xiang},
  journal={arXiv preprint arXiv:2403.04473},
  year={2024}
}

@article{hu2024mplug,
  title={mplug-docowl 1.5: Unified structure learning for ocr-free document understanding},
  author={Hu, Anwen and Xu, Haiyang and Ye, Jiabo and Yan, Ming and Zhang, Liang and Zhang, Bo and Li, Chen and Zhang, Ji and Jin, Qin and Huang, Fei and others},
  journal={arXiv preprint arXiv:2403.12895},
  year={2024}
}

@article{liu2024hrvda,
  title={HRVDA: High-Resolution Visual Document Assistant},
  author={Liu, Chaohu and Yin, Kun and Cao, Haoyu and Jiang, Xinghua and Li, Xin and Liu, Yinsong and Jiang, Deqiang and Sun, Xing and Xu, Linli},
  journal={arXiv preprint arXiv:2404.06918},
  year={2024}
}

@inproceedings{li2024monkey,
  title={Monkey: Image resolution and text label are important things for large multi-modal models},
  author={Li, Zhang and Yang, Biao and Liu, Qiang and Ma, Zhiyin and Zhang, Shuo and Yang, Jingxu and Sun, Yabo and Liu, Yuliang and Bai, Xiang},
  booktitle={Proceedings of the IEEE/CVF Conference on Computer Vision and Pattern Recognition},
  pages={26763--26773},
  year={2024}
}

@article{lu2023mathvista,
  title={Mathvista: Evaluating mathematical reasoning of foundation models in visual contexts},
  author={Lu, Pan and Bansal, Hritik and Xia, Tony and Liu, Jiacheng and Li, Chunyuan and Hajishirzi, Hannaneh and Cheng, Hao and Chang, Kai-Wei and Galley, Michel and Gao, Jianfeng},
  journal={arXiv preprint arXiv:2310.02255},
  year={2023}
}

@article{yue2023mmmu,
  title={Mmmu: A massive multi-discipline multimodal understanding and reasoning benchmark for expert agi},
  author={Yue, Xiang and Ni, Yuansheng and Zhang, Kai and Zheng, Tianyu and Liu, Ruoqi and Zhang, Ge and Stevens, Samuel and Jiang, Dongfu and Ren, Weiming and Sun, Yuxuan and others},
  journal={arXiv preprint arXiv:2311.16502},
  year={2023}
}

@article{guo2025deepseek,
  title={Deepseek-r1: Incentivizing reasoning capability in llms via reinforcement learning},
  author={Guo, Daya and Yang, Dejian and Zhang, Haowei and Song, Junxiao and Zhang, Ruoyu and Xu, Runxin and Zhu, Qihao and Ma, Shirong and Wang, Peiyi and Bi, Xiao and others},
  journal={arXiv preprint arXiv:2501.12948},
  year={2025}
}

@article{team2023gemini,
  title={Gemini: a family of highly capable multimodal models},
  author={Team, Gemini and Anil, Rohan and Borgeaud, Sebastian and Alayrac, Jean-Baptiste and Yu, Jiahui and Soricut, Radu and Schalkwyk, Johan and Dai, Andrew M and Hauth, Anja and Millican, Katie and others},
  journal={arXiv preprint arXiv:2312.11805},
  year={2023}
}

@misc{qwq32b,
    title = {QwQ-32B: Embracing the Power of Reinforcement Learning},
    url = {https://qwenlm.github.io/blog/qwq-32b/},
    author = {Qwen Team},
    month = {March},
    year = {2025}
}

@article{yang2025r1onevision,
  title={R1-Onevision: Advancing Generalized Multimodal Reasoning through Cross-Modal Formalization},
  author={Yi Yang and Xiaoxuan He and Hongkun Pan and Xiyan Jiang and Yan Deng and Xingtao Yang and Haoyu Lu and Dacheng Yin and Fengyun Rao and Minfeng Zhu and Bo Zhang and Wei Chen},
  journal={arXiv preprint arXiv:2503.10615},
  year={2025},
}

@article{liu2025visual,
  title={Visual-rft: Visual reinforcement fine-tuning},
  author={Liu, Ziyu and Sun, Zeyi and Zang, Yuhang and Dong, Xiaoyi and Cao, Yuhang and Duan, Haodong and Lin, Dahua and Wang, Jiaqi},
  journal={arXiv preprint arXiv:2503.01785},
  year={2025}
}

@article{peng2025lmmr1,
  title={Lmm-r1: Empowering 3b lmms with strong reasoning abilities through two-stage rule-based rl},
  author={Peng, Yingzhe and Zhang, Gongrui and Zhang, Miaosen and You, Zhiyuan and Liu, Jie and Zhu, Qipeng and Yang, Kai and Xu, Xingzhong and Geng, Xin and Yang, Xu},
  journal={arXiv preprint arXiv:2503.07536},
  year={2025}
}

@article{wang2025visualprm,
  title={Visualprm: An effective process reward model for multimodal reasoning},
  author={Wang, Weiyun and Gao, Zhangwei and Chen, Lianjie and Chen, Zhe and Zhu, Jinguo and Zhao, Xiangyu and Liu, Yangzhou and Cao, Yue and Ye, Shenglong and Zhu, Xizhou and others},
  journal={arXiv preprint arXiv:2503.10291},
  year={2025}
}

@article{liu2025othink,
  title={OThink-MR1: Stimulating multimodal generalized reasoning capabilities through dynamic reinforcement learning},
  author={Liu, Zhiyuan and Zhang, Yuting and Liu, Feng and Zhang, Changwang and Sun, Ying and Wang, Jun},
  journal={arXiv preprint arXiv:2503.16081},
  year={2025}
}

@article{zhou2025r1,
  title={R1-Zero's" Aha Moment" in Visual Reasoning on a 2B Non-SFT Model},
  author={Zhou, Hengguang and Li, Xirui and Wang, Ruochen and Cheng, Minhao and Zhou, Tianyi and Hsieh, Cho-Jui},
  journal={arXiv preprint arXiv:2503.05132},
  year={2025}
}

@article{meng2025mm,
  title={MM-Eureka: Exploring Visual Aha Moment with Rule-based Large-scale Reinforcement Learning},
  author={Meng, Fanqing and Du, Lingxiao and Liu, Zongkai and Zhou, Zhixiang and Lu, Quanfeng and Fu, Daocheng and Shi, Botian and Wang, Wenhai and He, Junjun and Zhang, Kaipeng and others},
  journal={arXiv preprint arXiv:2503.07365},
  year={2025}
}

@misc{chen2025r1v,
  author       = {Chen, Liang and Li, Lei and Zhao, Haozhe and Song, Yifan and Vinci},
  title        = {R1-V: Reinforcing Super Generalization Ability in Vision-Language Models with Less Than \$3},
  howpublished = {\url{https://github.com/Deep-Agent/R1-V}},
  note         = {Accessed: 2025-02-02},
  year         = {2025}
}

@misc{shen2025vlmr1,
  author       = {Shen, Haozhan and Zhang, Zilun and Zhao, Kangjia and Zhang, Qianqian and Xu, Ruochen and Zhao, Tiancheng},
  title        = {VLM-R1: A stable and generalizable R1-style Large Vision-Language Model},
  howpublished = {\url{https://github.com/om-ai-lab/VLM-R1}},
  note         = {Accessed: 2025-02-15},
  year         = {2025}
}

@article{vl-rethinker,
  title={VL-Rethinker: Incentivizing Self-Reflection of Vision-Language Models with Reinforcement Learning},
  author={Wang, Haozhe and Qu, Chao and Huang, Zuming and Chu, Wei and Lin, Fangzhen and Chen, Wenhu},
  journal={arXiv preprint arXiv:2504.08837},
  year={2025}
}

@inproceedings{zhang2024mathverse,
  title={Mathverse: Does your multi-modal llm truly see the diagrams in visual math problems?},
  author={Zhang, Renrui and Jiang, Dongzhi and Zhang, Yichi and Lin, Haokun and Guo, Ziyu and Qiu, Pengshuo and Zhou, Aojun and Lu, Pan and Chang, Kai-Wei and Qiao, Yu and others},
  booktitle={European Conference on Computer Vision},
  pages={169--186},
  year={2024},
  organization={Springer}
}

@article{wang2024measuring,
  title={Measuring multimodal mathematical reasoning with math-vision dataset},
  author={Wang, Ke and Pan, Junting and Shi, Weikang and Lu, Zimu and Ren, Houxing and Zhou, Aojun and Zhan, Mingjie and Li, Hongsheng},
  journal={Advances in Neural Information Processing Systems},
  volume={37},
  pages={95095--95169},
  year={2024}
}

@article{he2024olympiadbench,
  title={Olympiadbench: A challenging benchmark for promoting agi with olympiad-level bilingual multimodal scientific problems},
  author={He, Chaoqun and Luo, Renjie and Bai, Yuzhuo and Hu, Shengding and Thai, Zhen Leng and Shen, Junhao and Hu, Jinyi and Han, Xu and Huang, Yujie and Zhang, Yuxiang and others},
  journal={arXiv preprint arXiv:2402.14008},
  year={2024}
}

@article{liu2025noisyrollout,
  title={NoisyRollout: Reinforcing Visual Reasoning with Data Augmentation},
  author={Liu, Xiangyan and Ni, Jinjie and Wu, Zijian and Du, Chao and Dou, Longxu and Wang, Haonan and Pang, Tianyu and Shieh, Michael Qizhe},
  journal={arXiv preprint arXiv:2504.13055},
  year={2025}
}

@article{liu2024ocrbench,
  title={OCRBench: on the hidden mystery of OCR in large multimodal models},
  author={Liu, Yuliang and Li, Zhang and Huang, Mingxin and Yang, Biao and Yu, Wenwen and Li, Chunyuan and Yin, Xu-Cheng and Liu, Cheng-Lin and Jin, Lianwen and Bai, Xiang},
  journal={Science China Information Sciences},
  volume={67},
  number={12},
  pages={220102},
  year={2024},
  publisher={Springer}
}

@article{xu2025visulogic,
  title={VisuLogic: A Benchmark for Evaluating Visual Reasoning in Multi-modal Large Language Models},
  author={Xu, Weiye and Wang, Jiahao and Wang, Weiyun and Chen, Zhe and Zhou, Wengang and Yang, Aijun and Lu, Lewei and Li, Houqiang and Wang, Xiaohua and Zhu, Xizhou and others},
  journal={arXiv preprint arXiv:2504.15279},
  year={2025}
}

@inproceedings{johnson2017clevr,
  title={Clevr: A diagnostic dataset for compositional language and elementary visual reasoning},
  author={Johnson, Justin and Hariharan, Bharath and Van Der Maaten, Laurens and Fei-Fei, Li and Lawrence Zitnick, C and Girshick, Ross},
  booktitle={Proceedings of the IEEE conference on computer vision and pattern recognition},
  pages={2901--2910},
  year={2017}
}

@article{jaech2024openai,
  title={Openai o1 system card},
  author={Jaech, Aaron and Kalai, Adam and Lerer, Adam and Richardson, Adam and El-Kishky, Ahmed and Low, Aiden and Helyar, Alec and Madry, Aleksander and Beutel, Alex and Carney, Alex and others},
  journal={arXiv preprint arXiv:2412.16720},
  year={2024}
}

@article{cui2025process,
  title={Process reinforcement through implicit rewards},
  author={Cui, Ganqu and Yuan, Lifan and Wang, Zefan and Wang, Hanbin and Li, Wendi and He, Bingxiang and Fan, Yuchen and Yu, Tianyu and Xu, Qixin and Chen, Weize and others},
  journal={arXiv preprint arXiv:2502.01456},
  year={2025}
}

@inproceedings{singh2019towards,
  title={Towards vqa models that can read},
  author={Singh, Amanpreet and Natarajan, Vivek and Shah, Meet and Jiang, Yu and Chen, Xinlei and Batra, Dhruv and Parikh, Devi and Rohrbach, Marcus},
  booktitle={Proceedings of the IEEE/CVF conference on computer vision and pattern recognition},
  pages={8317--8326},
  year={2019}
}

@inproceedings{biten2019scene,
  title={Scene text visual question answering},
  author={Biten, Ali Furkan and Tito, Ruben and Mafla, Andres and Gomez, Lluis and Rusinol, Mar{\c{c}}al and Valveny, Ernest and Jawahar, CV and Karatzas, Dimosthenis},
  booktitle={Proceedings of the IEEE/CVF international conference on computer vision},
  pages={4291--4301},
  year={2019}
}

@article{fu2024ocrbench,
  title={OCRBench v2: An Improved Benchmark for Evaluating Large Multimodal Models on Visual Text Localization and Reasoning},
  author={Fu, Ling and Yang, Biao and Kuang, Zhebin and Song, Jiajun and Li, Yuzhe and Zhu, Linghao and Luo, Qidi and Wang, Xinyu and Lu, Hao and Huang, Mingxin and others},
  journal={arXiv preprint arXiv:2501.00321},
  year={2024}
}

@article{wadhawan2024contextual,
  title={Contextual: Evaluating context-sensitive text-rich visual reasoning in large multimodal models},
  author={Wadhawan, Rohan and Bansal, Hritik and Chang, Kai-Wei and Peng, Nanyun},
  journal={arXiv preprint arXiv:2401.13311},
  year={2024}
}

@article{li2024seed,
  title={Seed-bench-2-plus: Benchmarking multimodal large language models with text-rich visual comprehension},
  author={Li, Bohao and Ge, Yuying and Chen, Yi and Ge, Yixiao and Zhang, Ruimao and Shan, Ying},
  journal={arXiv preprint arXiv:2404.16790},
  year={2024}
}

@inproceedings{liu2024mmc,
  title={MMC: Advancing Multimodal Chart Understanding with Large-scale Instruction Tuning},
  author={Liu, Fuxiao and Wang, Xiaoyang and Yao, Wenlin and Chen, Jianshu and Song, Kaiqiang and Cho, Sangwoo and Yacoob, Yaser and Yu, Dong},
  booktitle={Proceedings of the 2024 Conference of the North American Chapter of the Association for Computational Linguistics: Human Language Technologies (Volume 1: Long Papers)},
  pages={1287--1310},
  year={2024}
}

@article{liu2024focus,
  title={Focus anywhere for fine-grained multi-page document understanding},
  author={Liu, Chenglong and Wei, Haoran and Chen, Jinyue and Kong, Lingyu and Ge, Zheng and Zhu, Zining and Zhao, Liang and Sun, Jianjian and Han, Chunrui and Zhang, Xiangyu},
  journal={arXiv preprint arXiv:2405.14295},
  year={2024}
}

@article{ouyang2024omnidocbench,
  title={Omnidocbench: Benchmarking diverse pdf document parsing with comprehensive annotations},
  author={Ouyang, Linke and Qu, Yuan and Zhou, Hongbin and Zhu, Jiawei and Zhang, Rui and Lin, Qunshu and Wang, Bin and Zhao, Zhiyuan and Jiang, Man and Zhao, Xiaomeng and others},
  journal={arXiv preprint arXiv:2412.07626},
  year={2024}
}

@article{wang2024charxiv,
  title={Charxiv: Charting gaps in realistic chart understanding in multimodal llms},
  author={Wang, Zirui and Xia, Mengzhou and He, Luxi and Chen, Howard and Liu, Yitao and Zhu, Richard and Liang, Kaiqu and Wu, Xindi and Liu, Haotian and Malladi, Sadhika and others},
  journal={Advances in Neural Information Processing Systems},
  volume={37},
  pages={113569--113697},
  year={2024}
}

@article{sun2021spatial,
  title={Spatial dual-modality graph reasoning for key information extraction},
  author={Sun, Hongbin and Kuang, Zhanghui and Yue, Xiaoyu and Lin, Chenhao and Zhang, Wayne},
  journal={arXiv preprint arXiv:2103.14470},
  year={2021}
}

@article{gan2024mme,
  title={MME-Finance: A Multimodal Finance Benchmark for Expert-level Understanding and Reasoning},
  author={Gan, Ziliang and Lu, Yu and Zhang, Dong and Li, Haohan and Liu, Che and Liu, Jian and Liu, Ji and Wu, Haipang and Fu, Chaoyou and Xu, Zenglin and others},
  journal={arXiv preprint arXiv:2411.03314},
  year={2024}
}

@article{bi2025verify,
  title={VERIFY: A Benchmark of Visual Explanation and Reasoning for Investigating Multimodal Reasoning Fidelity},
  author={Bi, Jing and Guo, Junjia and Liang, Susan and Sun, Guangyu and Song, Luchuan and Tang, Yunlong and He, Jinxi and Wu, Jiarui and Vosoughi, Ali and Chen, Chen and others},
  journal={arXiv preprint arXiv:2503.11557},
  year={2025}
}

@article{yang2024cc,
  title={CC-OCR: A Comprehensive and Challenging OCR Benchmark for Evaluating Large Multimodal Models in Literacy},
  author={Yang, Zhibo and Tang, Jun and Li, Zhaohai and Wang, Pengfei and Wan, Jianqiang and Zhong, Humen and Liu, Xuejing and Yang, Mingkun and Wang, Peng and Liu, Yuliang and others},
  journal={arXiv preprint arXiv:2412.02210},
  year={2024}
}

@article{zheng2023judging,
  title={Judging llm-as-a-judge with mt-bench and chatbot arena},
  author={Zheng, Lianmin and Chiang, Wei-Lin and Sheng, Ying and Zhuang, Siyuan and Wu, Zhanghao and Zhuang, Yonghao and Lin, Zi and Li, Zhuohan and Li, Dacheng and Xing, Eric and others},
  journal={Advances in Neural Information Processing Systems},
  volume={36},
  pages={46595--46623},
  year={2023}
}

@article{chang2024survey,
  title={A survey on evaluation of large language models},
  author={Chang, Yupeng and Wang, Xu and Wang, Jindong and Wu, Yuan and Yang, Linyi and Zhu, Kaijie and Chen, Hao and Yi, Xiaoyuan and Wang, Cunxiang and Wang, Yidong and others},
  journal={ACM transactions on intelligent systems and technology},
  volume={15},
  number={3},
  pages={1--45},
  year={2024},
  publisher={ACM New York, NY}
}

@article{qwen2.5,
    title   = {Qwen2.5 Technical Report}, 
    author  = {An Yang and Baosong Yang and Beichen Zhang and Binyuan Hui and Bo Zheng and Bowen Yu and Chengyuan Li and Dayiheng Liu and Fei Huang and Haoran Wei and Huan Lin and Jian Yang and Jianhong Tu and Jianwei Zhang and Jianxin Yang and Jiaxi Yang and Jingren Zhou and Junyang Lin and Kai Dang and Keming Lu and Keqin Bao and Kexin Yang and Le Yu and Mei Li and Mingfeng Xue and Pei Zhang and Qin Zhu and Rui Men and Runji Lin and Tianhao Li and Tingyu Xia and Xingzhang Ren and Xuancheng Ren and Yang Fan and Yang Su and Yichang Zhang and Yu Wan and Yuqiong Liu and Zeyu Cui and Zhenru Zhang and Zihan Qiu},
    journal = {arXiv preprint arXiv:2412.15115},
    year    = {2024}
}

@article{bai2025qwen2,
  title={Qwen2. 5-vl technical report},
  author={Bai, Shuai and Chen, Keqin and Liu, Xuejing and Wang, Jialin and Ge, Wenbin and Song, Sibo and Dang, Kai and Wang, Peng and Wang, Shijie and Tang, Jun and others},
  journal={arXiv preprint arXiv:2502.13923},
  year={2025}
}

@misc{kimiteam2025kimivltechnicalreport,
      title={{Kimi-VL} Technical Report}, 
      author={Kimi Team and Angang Du and Bohong Yin and Bowei Xing and Bowen Qu and Bowen Wang and Cheng Chen and Chenlin Zhang and Chenzhuang Du and Chu Wei and Congcong Wang and Dehao Zhang and Dikang Du and Dongliang Wang and Enming Yuan and Enzhe Lu and Fang Li and Flood Sung and Guangda Wei and Guokun Lai and Han Zhu and Hao Ding and Hao Hu and Hao Yang and Hao Zhang and Haoning Wu and Haotian Yao and Haoyu Lu and Heng Wang and Hongcheng Gao and Huabin Zheng and Jiaming Li and Jianlin Su and Jianzhou Wang and Jiaqi Deng and Jiezhong Qiu and Jin Xie and Jinhong Wang and Jingyuan Liu and Junjie Yan and Kun Ouyang and Liang Chen and Lin Sui and Longhui Yu and Mengfan Dong and Mengnan Dong and Nuo Xu and Pengyu Cheng and Qizheng Gu and Runjie Zhou and Shaowei Liu and Sihan Cao and Tao Yu and Tianhui Song and Tongtong Bai and Wei Song and Weiran He and Weixiao Huang and Weixin Xu and Xiaokun Yuan and Xingcheng Yao and Xingzhe Wu and Xinxing Zu and Xinyu Zhou and Xinyuan Wang and Y. Charles and Yan Zhong and Yang Li and Yangyang Hu and Yanru Chen and Yejie Wang and Yibo Liu and Yibo Miao and Yidao Qin and Yimin Chen and Yiping Bao and Yiqin Wang and Yongsheng Kang and Yuanxin Liu and Yulun Du and Yuxin Wu and Yuzhi Wang and Yuzi Yan and Zaida Zhou and Zhaowei Li and Zhejun Jiang and Zheng Zhang and Zhilin Yang and Zhiqi Huang and Zihao Huang and Zijia Zhao and Ziwei Chen},
      year={2025},
      eprint={2504.07491},
      archivePrefix={arXiv},
      primaryClass={cs.CV},
      url={https://arxiv.org/abs/2504.07491}, 
}

@article{meta2025llama,
  title={The Llama 4 herd: The beginning of a new era of natively multimodal AI innovation},
  author={Meta, AI},
  journal={https://ai. meta. com/blog/llama-4-multimodal-intelligence/, checked on},
  volume={4},
  number={7},
  pages={2025},
  year={2025}
}

@article{anthropic2025claude37,
  title={Claude 3.7 Sonnet Extended Thinking},
  author={Anthropic},
  journal={Anthropic System Card},
  year={2025},
  url={https://assets.anthropic.com/m/785e231869ea8b3b/original/claude-3-7-sonnet-system-card.pdf},
}

@article{deepmind2025flashthinking,
  title={Gemini 2.0 Flash Thinking},
  author={DeepMind},
  journal={Google DeepMind website},
  year={2025},
  url={https://deepmind.google/technologies/gemini/flash-thinking/},
}

@article{hurst2024gpt,
  title={Gpt-4o system card},
  author={Hurst, Aaron and Lerer, Adam and Goucher, Adam P and Perelman, Adam and Ramesh, Aditya and Clark, Aidan and Ostrow, AJ and Welihinda, Akila and Hayes, Alan and Radford, Alec and others},
  journal={arXiv preprint arXiv:2410.21276},
  year={2024}
}

@misc{chen2025sftrlearlyinvestigation,
      title={SFT or RL? An Early Investigation into Training R1-Like Reasoning Large Vision-Language Models}, 
      author={Hardy Chen and Haoqin Tu and Fali Wang and Hui Liu and Xianfeng Tang and Xinya Du and Yuyin Zhou and Cihang Xie},
      year={2025},
      eprint={2504.11468},
      archivePrefix={arXiv},
      primaryClass={cs.CL},
      url={https://arxiv.org/abs/2504.11468}, 
}

@misc{li2024llmsasjudgescomprehensivesurveyllmbased,
      title={LLMs-as-Judges: A Comprehensive Survey on LLM-based Evaluation Methods}, 
      author={Haitao Li and Qian Dong and Junjie Chen and Huixue Su and Yujia Zhou and Qingyao Ai and Ziyi Ye and Yiqun Liu},
      year={2024},
      eprint={2412.05579},
      archivePrefix={arXiv},
      primaryClass={cs.CL},
      url={https://arxiv.org/abs/2412.05579}, 
}

@article{zhu2025internvl3,
  title={InternVL3: Exploring Advanced Training and Test-Time Recipes for Open-Source Multimodal Models},
  author={Zhu, Jinguo and Wang, Weiyun and Chen, Zhe and Liu, Zhaoyang and Ye, Shenglong and Gu, Lixin and Duan, Yuchen and Tian, Hao and Su, Weijie and Shao, Jie and others},
  journal={arXiv preprint arXiv:2504.10479},
  year={2025}
}

@article{guo2025seed1,
  title={Seed1. 5-VL Technical Report},
  author={Guo, Dong and Wu, Faming and Zhu, Feida and Leng, Fuxing and Shi, Guang and Chen, Haobin and Fan, Haoqi and Wang, Jian and Jiang, Jianyu and Wang, Jiawei and others},
  journal={arXiv preprint arXiv:2505.07062},
  year={2025}
}

@article{li2022pp,
  title={PP-OCRv3: More attempts for the improvement of ultra lightweight OCR system},
  author={Li, Chenxia and Liu, Weiwei and Guo, Ruoyu and Yin, Xiaoting and Jiang, Kaitao and Du, Yongkun and Du, Yuning and Zhu, Lingfeng and Lai, Baohua and Hu, Xiaoguang and others},
  journal={arXiv preprint arXiv:2206.03001},
  year={2022}
}

@misc{o3mini,
  author       = {{Open AI Team}},
  title        = {OpenAI o3-mini},
  howpublished = {\url{https://openai.com/index/openai-o3-mini}},
  note         = {Accessed: 2025-01-31},
  year         = {2025}
}

@inproceedings{huang2024mini,
  title={Mini-monkey: Alleviating the semantic sawtooth effect for lightweight mllms via complementary image pyramid},
  author={Huang, Mingxin and Liu, Yuliang and Liang, Dingkang and Jin, Lianwen and Bai, Xiang},
  booktitle={International Conference on Learning Representations}
}

@article{hu2024mplug15,
  title={mplug-docowl 1.5: Unified structure learning for ocr-free document understanding},
  author={Hu, Anwen and Xu, Haiyang and Ye, Jiabo and Yan, Ming and Zhang, Liang and Zhang, Bo and Li, Chen and Zhang, Ji and Jin, Qin and Huang, Fei and others},
  journal={arXiv preprint arXiv:2403.12895},
  year={2024}
}

@article{guan2025token,
  title={A Token-level Text Image Foundation Model for Document Understanding},
  author={Guan, Tongkun and Wang, Zining and Fu, Pei and Guo, Zhengtao and Shen, Wei and Zhou, Kai and Yue, Tiezhu and Duan, Chen and Sun, Hao and Jiang, Qianyi and others},
  journal={arXiv preprint arXiv:2503.02304},
  year={2025}
}

@article{yu2024texthawk2,
  title={Texthawk2: A large vision-language model excels in bilingual ocr and grounding with 16x fewer tokens},
  author={Yu, Ya-Qi and Liao, Minghui and Zhang, Jiwen and Wu, Jihao},
  journal={arXiv preprint arXiv:2410.05261},
  year={2024}
}

@inproceedings{zhang2025dockylin,
  title={Dockylin: A large multimodal model for visual document understanding with efficient visual slimming},
  author={Zhang, Jiaxin and Yang, Wentao and Lai, Songxuan and Xie, Zecheng and Jin, Lianwen},
  booktitle={Proceedings of the AAAI Conference on Artificial Intelligence},
  volume={39},
  number={9},
  pages={9923--9932},
  year={2025}
}

@article{team2025kwai,
  title={Kwai Keye-VL Technical Report},
  author={Team, Kwai Keye and Yang, Biao and Wen, Bin and Liu, Changyi and Chu, Chenglong and Song, Chengru and Rao, Chongling and Yi, Chuan and Li, Da and Zang, Dunju and others},
  journal={arXiv preprint arXiv:2507.01949},
  year={2025}
}

@misc{coreteam2025mimovltechnicalreport,
      title={MiMo-VL Technical Report}, 
      author={LLM-Core-Team Xiaomi},
      year={2025},
      eprint={2506.03569},
      archivePrefix={arXiv},
      primaryClass={cs.CL},
      url={https://arxiv.org/abs/2506.03569}, 
}

@article{hong2025glm,
  title={Glm-4.1 v-thinking: Towards versatile multimodal reasoning with scalable reinforcement learning},
  author={Hong, Wenyi and Yu, Wenmeng and Gu, Xiaotao and Wang, Guo and Gan, Guobing and Tang, Haomiao and Cheng, Jiale and Qi, Ji and Ji, Junhui and Pan, Lihang and others},
  journal={arXiv e-prints},
  pages={arXiv--2507},
  year={2025}
}

@article{Qwen2-VL,
  title={Qwen2-VL: Enhancing Vision-Language Model's Perception of the World at Any Resolution},
  author={Wang, Peng and Bai, Shuai and Tan, Sinan and Wang, Shijie and Fan, Zhihao and Bai, Jinze and Chen, Keqin and Liu, Xuejing and Wang, Jialin and Ge, Wenbin and Fan, Yang and Dang, Kai and Du, Mengfei and Ren, Xuancheng and Men, Rui and Liu, Dayiheng and Zhou, Chang and Zhou, Jingren and Lin, Junyang},
  journal={arXiv preprint arXiv:2409.12191},
  year={2024}
}

@inproceedings{hu2024mplug25,
    title = "m{PLUG}-{D}oc{O}wl2: High-resolution Compressing for {OCR}-free Multi-page Document Understanding",
    author = "Hu, Anwen  and
      Xu, Haiyang  and
      Zhang, Liang  and
      Ye, Jiabo  and
      Yan, Ming  and
      Zhang, Ji  and
      Jin, Qin  and
      Huang, Fei  and
      Zhou, Jingren",
    booktitle = "Proceedings of the 63rd Annual Meeting of the Association for Computational Linguistics (Volume 1: Long Papers)",
    year = "2025",
    pages = "5817--5834",
}

@inproceedings{duan2025docopilot,
  title={Docopilot: Improving Multimodal Models for Document-Level Understanding},
  author={Duan, Yuchen and Chen, Zhe and Hu, Yusong and Wang, Weiyun and Ye, Shenglong and Shi, Botian and Lu, Lewei and Hou, Qibin and Lu, Tong and Li, Hongsheng and others},
  booktitle={Proceedings of the Computer Vision and Pattern Recognition Conference},
  pages={4026--4037},
  year={2025}
}

@inproceedings{xiao2025adaptive,
  title={Adaptive Markup Language Generation for Contextually-Grounded Visual Document Understanding},
  author={Xiao, Han and Xie, Yina and Tan, Guanxin and Chen, Yinghao and Hu, Rui and Wang, Ke and Zhou, Aojun and Li, Hao and Shao, Hao and Lu, Xudong and others},
  booktitle={Proceedings of the Computer Vision and Pattern Recognition Conference},
  pages={29558--29568},
  year={2025}
}

@inproceedings{wang2025marten,
  title={Marten: Visual question answering with mask generation for multi-modal document understanding},
  author={Wang, Zining and Guan, Tongkun and Fu, Pei and Duan, Chen and Jiang, Qianyi and Guo, Zhentao and Guo, Shan and Luo, Junfeng and Shen, Wei and Yang, Xiaokang},
  booktitle={Proceedings of the Computer Vision and Pattern Recognition Conference},
  pages={14460--14471},
  year={2025}
}

@article{zhang2025thyme,
  title={Thyme: Think Beyond Images},
  author={Zhang, Yi-Fan and Lu, Xingyu and Yin, Shukang and Fu, Chaoyou and Chen, Wei and Hu, Xiao and Wen, Bin and Jiang, Kaiyu and Liu, Changyi and Zhang, Tianke and others},
  journal={arXiv preprint arXiv:2508.11630},
  year={2025}
}

@article{hong2025deepeyesv2,
  title={DeepEyesV2: Toward Agentic Multimodal Model},
  author={Hong, Jack and Zhao, Chenxiao and Zhu, ChengLin and Lu, Weiheng and Xu, Guohai and Yu, Xing},
  journal={arXiv preprint arXiv:2511.05271},
  year={2025}
}
